\documentclass[acmsmall]{acmart}

\AtBeginDocument{%
  }

\usepackage{threeparttable}
\usepackage{multirow}
\usepackage{subfigure}
\usepackage{array}
\usepackage{bigstrut}
\usepackage{xcolor}
\usepackage{enumitem}
\usepackage{colortbl}
\definecolor{Ocean}{RGB}{200, 230, 250}

\acmJournal{JACM}
\acmVolume{37}
\acmNumber{4}
\acmArticle{111}
\acmMonth{8}

\begin{document}

\title{CAT-LLM: Style-enhanced Large Language Models with Text Style Definition for Chinese Article-style Transfer}

\author{Zhen Tao}
\affiliation{%
  \institution{Renmin University of China}
  \city{Beijing}
  \country{China}
}
\email{taozhen@ruc.edu.cn}

\author{Dinghao Xi}
\authornote{Corresponding authors.}
\affiliation{%
  \institution{Shanghai University of Finance and Economics}
  \city{Shanghai}
  % \state{Texas}
  \country{China}}
\email{xidinghao@mail.shufe.edu.cn}

\author{Zhiyu Li}
\authornotemark[1]
\affiliation{%
  \institution{Institute for Advanced Algorithms Research}
  \city{Shanghai}
  \country{China}
}
\email{lizy@iaar.ac.cn}

\author{Liumin Tang}
\affiliation{%
  \institution{Renmin University of China}
  \city{Beijing}
  \country{China}}
\email{tangliumin@ruc.edu.cn}

\author{Wei Xu}
\affiliation{%
  \institution{Renmin University of China}
  \city{Beijing}
  \country{China}}
\email{weixu@ruc.edu.cn}

\begin{abstract}
Text style transfer plays a vital role in online entertainment and social media. However, existing models struggle to handle the complexity of Chinese long texts, such as rhetoric, structure, and culture, which restricts their broader application. To bridge this gap, we propose a \textbf{C}hinese \textbf{A}rticle-style \textbf{T}ransfer (\textbf{CAT-LLM}) framework, which addresses the challenges of style transfer in complex Chinese long texts. At its core, CAT-LLM features a bespoke pluggable \textbf{T}ext \textbf{S}tyle \textbf{D}efinition (\textbf{TSD}) module that integrates machine learning algorithms to analyze and model article styles at both word and sentence levels. This module acts as a bridge, enabling large language models (LLMs) to better understand and adapt to the complexities of Chinese article styles. Furthermore, it supports the dynamic expansion of internal style trees, enabling the framework to seamlessly incorporate new and diverse style definitions, enhancing adaptability and scalability for future research and applications. Additionally, to facilitate robust evaluation, we created ten parallel datasets using a combination of ChatGPT and various Chinese texts, each corresponding to distinct writing styles, significantly improving the accuracy of the model evaluation and establishing a novel paradigm for text style transfer research. Extensive experimental results demonstrate that CAT-LLM, combined with GPT-3.5-Turbo, achieves state-of-the-art performance, with a transfer accuracy F1 score of 79.36\% and a content preservation F1 score of 96.47\% on the ``Fortress Besieged'' dataset. These results highlight CAT-LLM's innovative contributions to style transfer research, including its ability to preserve content integrity while achieving precise and flexible style transfer across diverse Chinese text domains. Building on these contributions, CAT-LLM presents significant potential for advancing Chinese digital media and facilitating automated content creation. Source code is available at GitHub\footnote{https://github.com/TaoZhen1110/CAT-LLM}.
\end{abstract}




% \received{30 June 2024}
% \received[revised]{25 January 2025}
% \received[accepted]{2 June 2025}

\maketitle

\section{Introduction}

Text style transfer is a pivotal advancement within the domain of Natural Language Processing (NLP), serving as a bridge between content fidelity and stylistic transformation. The primary objective of this technique is to modify the stylistic attributes of a given text while meticulously preserving its original semantic content, enabling a seamless transition between different stylistic renditions of the same information \cite{toshevska2021, yi2021}. The applications of text style transfer are diverse, with notable progress in areas such as fake news detection \cite{przybyla2020}, where it identifies stylistic inconsistencies, and social media \cite{wang2023rolellm}, where it enhances user engagement by tailoring content to audience preferences. In addition, it plays a key role in cultural preservation by modernizing classical texts while retaining their stylistic essence \cite{SC2acl2024}, and in customer communication, improving businesses to adapt messaging for different audiences to improve user experience \cite{kushwaha2024markbot}.

\begin{figure}[b]
  \centering
  \includegraphics[width=\textwidth]{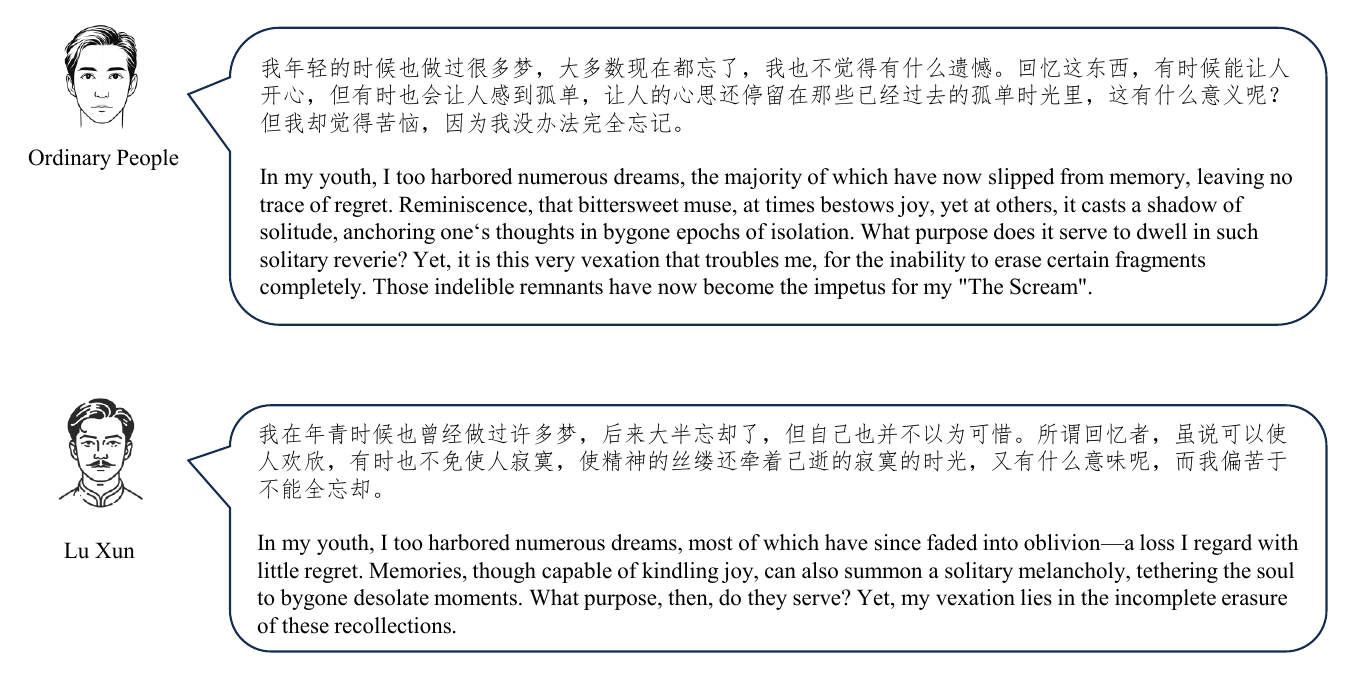}
   \caption{An example that transfer styleless text to ``The Scream'' style text.}
   \label{fig1}
\end{figure}

Before the advent of large language models (LLMs), the frameworks for text style transfer predominantly rely on small deep learning models. In accordance with the operational process of text transfer, these models can be broadly classified into two main categories: end-to-end models and two-stage models. Despite their innovative approaches, these earlier models often exhibited lower transfer accuracy and were primarily effective for relatively straightforward style transfer tasks, such as emotion \cite{yi2021}, politeness \cite{danescu2013}, and etiquette \cite{sheikha2010}. A significant limitation in previous research was the absence of large-scale parallel corpora, which necessitated the use of non-parallel datasets to train classifiers for evaluating the transfer accuracy of unsupervised models. This constraint led to notable evaluation errors, as an author's style is frequently tied to specific thematic elements. For instance, the term ``Camel Xiangzi'' predominantly appears in the work of Lao She, making accurate evaluation challenging without parallel datasets. Moreover, the majority of existing studies have focused on style transfer at the sentences level in English, with limited research dedicated to longer Chinese texts. Given the distinct linguistic structures, idiomatic expressions, and cultural nuances inherent in Chinese, there is a compelling need to design models specifically tailored for the style transfer of extended Chinese texts. Figure \ref{fig1} illustrates an example of Chinese article-style transfer, which transforms ordinary vernacular into a fragment of the article ``The Scream''.

Recently, LLMs have demonstrated the ability to handle more complex NLP tasks such as summarization generation and role-playing through zero-shot transfer without the need for dedicated task-specific training \cite{brown2020,song2023llm}, attracting significant attention from both academia and industry. Given the strong capabilities of LLMs in semantic understanding and text generation, they can effectively address challenges faced by small models of style transfer, such as the lack of parallel data and low transfer accuracy. Currently, several studies have made significant headway in text style transfer leveraging LLMs. Lai et al. \cite{lai2023} employ ChatGPT as a multidimensional evaluator, assessing the style strength, content preservation, and fluency of text generated by various small-scale text style transfer models. Shanahan \cite{shanahan2023role} and Wang et al. \cite{wang2023rolellm} investigate the use of LLMs for role-playing in conversational agents, where models like ChatGPT are tasked with generating text or dialogue in a specific role's style. A significant limitation of these methods is their difficulty in accurately mimicking the writing style of relatively unknown authors or anonymous works, often resulting in comprehension challenges and potential hallucinatory issues. In addition, the diverse and intricate expression methods in Chinese, such as idioms, rhetorical devices, and literary styles, pose significant challenges. Directly inputting works from a specified author into LLMs not only consumes more tokens, affecting inference speed, but also can lead to insufficient understanding of the author’s writing style if only text segments are input. Moreover, this approach often lacks interpretability and guidance, making it difficult to capture the rich and diverse expressions inherent in Chinese article styles. Without clear insights into how stylistic features are identified and transferred, the outputs risk deviating from the intended style, underscoring the need for more nuanced methods that ensure fidelity while enhancing understanding of the transformation process.

To overcome the above challenges, we propose a \textbf{C}hinese \textbf{A}rticle-style \textbf{T}ransfer (\textbf{CAT-LLM}) framework based on LLMs. At the heart of CAT-LLM is a pluggable \textbf{T}ext \textbf{S}tyle \textbf{D}efinition (\textbf{TSD}) module that utilizes machine learning algorithms to analyze and model article styles at both the word and sentence levels. This module not only enables large language models to better capture the intricacies of Chinese article styles but also supports the dynamic expansion of internal style trees, facilitating seamless integration of new and diverse style definitions. Leveraging this comprehensive style definition, we design style-enhanced prompts to guide LLMs in generating high-quality, style-consistent text. Furthermore, we constructed large parallel datasets using ChatGPT, generating styleless text from ten distinct types of Chinese long texts to enable a more precise performance evaluation. These datasets represent a significant step forward in establishing a robust benchmark for text style transfer research. Extensive experiments demonstrate that our framework surpasses state-of-the-art research in both style accuracy and content preservation. The primary contributions are summarized as follows:

\begin{itemize}
\item We propose CAT-LLM, a pioneering Chinese Article-style Transfer framework to address the complexities of style transfer in Chinese long texts. Extensive experiments demonstrate that CAT-LLM achieves state-of-the-art performance in style accuracy and content preservation.

\item We design a pluggable Text Style Definition (TSD) module that integrates various small models to analyze and model text styles at both the word and sentence levels. This module enhances LLMs' style transfer capabilities and allows dynamic expansion of style trees for seamless adaptation to diverse and evolving definitions.

\item We create ten parallel datasets covering distinct types of Chinese long texts to facilitate robust and precise evaluation of style transfer models. This contribution establishes a novel paradigm for evaluating text style transfer and paves the way for future research.
\end{itemize}

\section{Related Work}\label{sec2}

\subsection{Large Language Models}

The advent of Large Language Models (LLMs) has significantly transformed the landscape of Natural Language Processing (NLP), shifting the focus from traditional tasks like translation and classification to more sophisticated applications such as role-playing and dialogue systems \cite{ai4science2023impact, che2023towards}. This marks a substantial enhancement in the scope and capability of NLP technologies, allowing for more complex language comprehension and generation tasks. LLMs like Baichuan \cite{wen2023chathome} and ChatGLM \cite{hou2024chatglm} exemplify the cutting-edge applications of these technologies. Baichuan, developed to support a wide range of Chinese language tasks, excels in understanding and generating complex Chinese texts, making it a versatile tool for applications such as translation, sentiment analysis, and dialogue systems. It leverages extensive training on diverse datasets to provide high accuracy and contextual relevance, catering to the unique linguistic and cultural nuances of Chinese. ChatGLM is specifically designed for conversational AI, making it ideal for applications like customer service and virtual assistants. It generates human-like responses by ensuring coherence, relevance, and contextual accuracy, which enhances user engagement and satisfaction.

In terms of the openness of the LLMs source code, it can be broadly categorized into open-source and closed-source models \cite{liang2023uhgeval}. Open-source models, such as LLaMA and its variants, have made high-performing NLP models accessible to a wider range of researchers and developers. LLaMA models range from 7B to 70B parameters and are trained on vast datasets, demonstrating competitive performance across various benchmarks \cite{touvron2023llama}. Closed-source models, represented by OpenAI's GPT series, particularly GPT-4, push the boundaries of what LLMs can achieve. GPT-4, with over a trillion parameters, excels in diverse tasks including complex reasoning, coding, and multilingual understanding \cite{ai4science2023impact}. One of the critical innovations in leveraging LLMs is the design of effective prompts to guide their outputs. Shao et al. \cite{shao2023prompting} propose Prophet to prompt LLMs with answer heuristics for knowledge-based Visual Question Answering. Jang et al. \cite{jang2023can} investigate the effectiveness of negated prompts in directing LLMs, revealing limitations in their ability to follow such prompts accurately. Singh et al. \cite{singh2023} introduce a procedural LLMs prompt structure that enable LLMs to directly generate sequences of robot operations during task planning. Additionally, Li et al. \cite{li2023} propose a directed stimulus prompt to guide LLMs in achieving specific desired outputs. In summary, the evolution of LLMs has opened new frontiers in NLP research, from enabling complex agent-level tasks to refining the art of prompt engineering. These advancements not only enhance the practical applications of NLP but also pave the way for more robust and interpretable AI systems.

\subsection{Text Style Transfer}

Text style transfer has been a focus in NLP research, attracting widespread attention in computer science and literature. Hu et al. \cite{Hu} pioneer the introduction of the concept of style transfer from image to text, laying a cornerstone for subsequent explorations in text style transfer.

Recent advancements in NLP have led to the development of numerous text style transfer methods. These methods can be broadly categorized into two main families based on their transfer processes. The first family views content and style in a sentence as relatively independent elements, focusing on separating the original style from the content and replacing it with the target style. For example, Lee et al. \cite{Dongkyu-etal} proposed a two-stage method: first, extracting and removing attribute markers using a pre-trained classifier, and then employing a Transformer-based generator to combine the target attribute with content tokens. Similarly, Tian et al. \cite{Dongkyu-etal} utilized reverse attention to implicitly remove style information from each token, preserving the original content, and introduced conditional layer normalization to create content-dependent style representations. StoryTrans \cite{zhu-etal} marked a significant advance in transferring long-form text. It used discourse representations to capture source content information and transform it into the target style through learnable style embeddings. To further enhance content preservation, the model employed a mask-and-fill framework, incorporating style-specific keywords from the source text into the generation process. However, these methods face challenges in maintaining the integrity of the original textual content, as certain tokens within sentences and paragraphs can embody both content and distinct stylistic characteristics. The second family focuses on designing end-to-end models for style transfer. The Style Transformer \cite{dai-etal} leverages the robust capabilities of the Transformer architecture to handle long-term dependencies in text, directly providing the target style embedding to the decoder without the need for explicit disentanglement. StyIns \cite{yi2021} employs a generative flow technique to construct a distinctive and expressive latent style space, delivering robust style signals to the attention-based decoder supported by multiple style instances. Lyu et al. \cite{lyu2023fine} proposed a diffusion-based language model capable of performing fine-grained text style transfer from scratch, utilizing a limited dataset without relying on external weights or tools. This model also supports multitasking and the composition of multiple fine-grained transfers. However, this family primarily addresses sentiment and tense transfer in individual English sentences, with limited research conducted on style transfer for longer Chinese texts. 

Building on the advancements discussed, our work proposes a novel text style transfer framework based on large language models, specifically targeting long Chinese texts. This approach addresses the shortcomings of existing methods in terms of content preservation and stylistic accuracy.

\section{Methodology}\label{sec3}

\subsection{Overview of Task and Framework}

To make our discussion clearer, we first define the task of Chinese article-style transfer. Suppose that there is a set of datasets $\{D_{i}\}_{i=1}^{K}$, each containing numerous textual fragments and sharing the distinctive style $\{v_{i}\}_{i=1}^{K}$. The task is to rewrite a article fragment $x$ into a new fragment $x^{'}$ with the target style $v^{'}$. The process ensure that the new fragment $x^{'}$ possesses the target style $v^{'}$ while preserving as much information from the original fragment as possible. 

\begin{figure}[t]
  \centering
  \includegraphics[width=\textwidth]{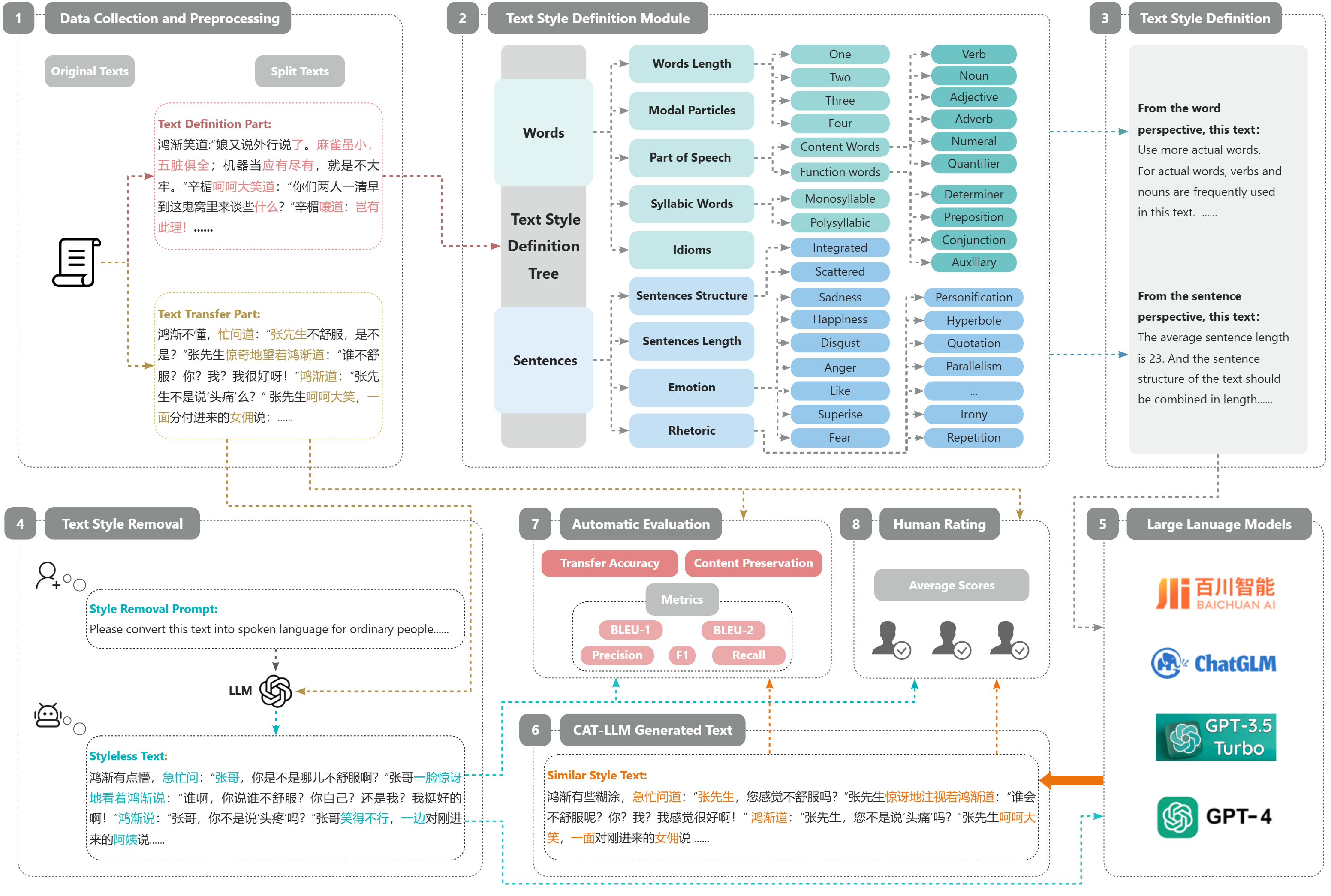}
   \caption{Illustration of the proposed CAT-LLM framework. For the examples of Text Transfer Part, Styleless Text, and Similar Style Text in the figure, we have highlighted phrases that represent text styles with different colors to facilitate a better comparison of the text style transfer process.}
   \label{fig2}
\end{figure}

Our proposed Chinese Article-style Transfer (CAT-LLM) framework is divided into eight stages. The specific structure of the framework is shown in Figure \ref{fig2}. In the first stage, we divide a certain corpus $D_{i}$ into two parts: the style definition part $D_{i1}$ and the style transfer part $D_{i2}$. In the second stage, we propose a Text Style Definition (TSD) module, which comprehensively considers style features based on the style definition part $D_{i1}$ from the word and sentence levels. Given that the text acquisition in the style definition part is relatively random and extensive, we consider the summarized style features as the overall style features of the article. In the third stage, based on the text features extracted by the TSD module, we construct a comprehensive structured textual representation that encompasses features at both the word and sentence levels. In the fourth stage, we transform $D_{i2}$ into styleless text $D_{i2}^{'}$ using ChatGPT to construct the parallel datasets. This strategy addresses issues such as missing labels and evaluation difficulties in unsupervised learning, providing an innovative approach to dataset construction in the era of large language models. In the fifth and sixth stages, we construct a style-enhanced prompt to guide LLMs in generating new text that is consistent with the original article style, while retaining styleless text information. In the final stages, newly generated texts $D_{i2}^{''}$ are compared with the original style transfer part $D_{i2}$ and the corresponding styleless text $D_{i2}^{'}$ to comprehensively assess the performance of the style transfer framework, using both automatic evaluation and human rating. Our research primarily centers on TSD module design and prompt construction, which will be elaborated upon in the subsequent text.

\subsection{Text Style Definition Module}

As shown in Figure \ref{fig2}, we propose the Text Style Definition (TSD) module, which computes the style of the style definition part $D_{i1}$ at both the word and sentence levels, providing precise style definitions. The TSD module is designed to enhance the accurate identification of text styles and to provide a clear and comprehensive prompt of styles for large language models (LLMs). Additionally, this module can be seamlessly integrated into various LLMs to optimize their performance in style transfer. Moreover, it offers interpretability for researchers, enabling them to gain deeper insights into the underlying mechanisms of text style recognition and transfer within the models.

\subsubsection{\textbf{Word Level}} 

Words form the cornerstone of constructing text and play a crucial role in shaping language features \cite{Serrano2009Modeling,Castillo2018The}. Quantitative analysis of the use of vocabulary in texts can reveal language style differences between different texts. In this section, we conduct an in-depth analysis of the words style of the article from multiple perspectives such as part of speech, words length and syllabic words, modal particles and idioms.

\textbf{Part of Speech.} 
To analyze the part-of-speech (POS) distribution in the text $D_{i2}$, we employ the $Cut$ function from the $jieba$\footnote{https://github.com/fxsjy/jieba} library to tokenize the text and obtain POS annotations for each word. Let $W=(w_{1},w_{2},...,w_{i},w_{n_{w}})$ represent the sequence of the words, where $n_{w}$ is the total number of words, and $T=(t_{1},t_{2},...,t_{i},t_{n_{w}})$ denote the corresponding sequence of POS tags. The POS tags are primarily categorized into content words, including nouns, verbs, and adjectives, and function words, such as adverbs, prepositions, and conjunctions. Through classification and statistical analysis of the vocabulary within an article, we can infer the distribution of part-of-speech categories.

To identify the predominant part-of-speech style of the article, we define the function $I_{ps}(\cdot)$ to count the occurrences of each POS type based on the tags $T$. The POS style $f_{ps}$ of the article is determined by selecting the POS type with the highest frequency. This process is formalized as follows:
\begin{equation}
f_{ps} = \underset{i\in \{1,2,...,n_{w}\}}{\rm argmax}\{I_{ps}(w_{i},t_{i})\},
\end{equation}
where the function $I_{ps}(w_{i},t_{i})$ is defined as:
\begin{equation}
I_{ps}(w_i, t_i) = \sum_{j=1}^{n_{w}} \delta(t_j - t_i) \cdot \mathbb{1}(w_j = w_i),
\end{equation}
and $\delta(t_j - t_i)$ is the Kronecker delta function which equals 1 if \(t_j = t_i\) and 0 otherwise, and \(\mathbb{1}(\cdot)\) is an indicator function that returns 1 when its argument is true and 0 otherwise. The goal is to maximize \(I_{ps}(w_i, t_i)\), thereby determining the POS type \(t_i\) with the highest relative frequency in the text.

\textbf{Word Length and Syllabic Words.} The selection of word lengths by authors often reflects their personal style and linguistic preferences \cite{Lewis2016The}. To elucidate the characteristic word length profile of a given text, we conduct a comprehensive statistical analysis of word lengths. This process is formalized as follows:
\begin{equation}
f_{l} = \underset{i\in \{1,2,...,n_{w}\}}{\rm argTop}K_{l} \{I_{l}(len(w_{i}))\},
\end{equation}
where $len(\cdot)$ is a function that computes the length of each word, and $I_{l}(\cdot)$ is an indicator function that enumerates the frequency of words with specific lengths. The top $K_{l}$ word length categories, determined by their frequency, are then selected to represent the predominant word length characteristics of the text. 

To further refine our analysis, we categorize words into monosyllabic and polysyllabic groups. We then identify the top $K_{mo}$ and top $K_{po}$ most frequently occurring words within these groups, respectively, to serve as representative syllabic word types. This is mathematically represented as:
\begin{equation}
f_{mo},f_{po} = {\rm argTop}K_{sy}\{freq(W_{mo},W_{po})\},
\end{equation}
where $W_{mo}$ and $W_{po}$ denote the sets of monosyllabic and polysyllabic words, respectively, and the $freq(\cdot)$ is a function that computes the occurrence frequency of each word within these sets. By examining these distributions, we infer the author's stylistic inclinations toward the use of monosyllabic and polysyllabic words.

\textbf{Modal Particles and idioms.} The analysis of modal particles and idioms provides insights into the linguistic style and expressive tendencies of the text. Utilizing established modal particle\footnote{https://baike.baidu.com/} and idiom databases\footnote{https://github.com/crazywhalecc/idiom-database}, we perform regular expression matching to extract the most frequently occurring modal particles and idioms. This method enables us to identify representative modal particles and idioms that characterize the text. The primary modal particles and idioms are formalized as follows: 
\begin{equation}
f_{modal} = \underset{i\in \{1,2,...,n_{w}\}}{\rm argTop}K_{modal}\{freq(Re(w_{i},D_{modal}))\},
\end{equation}
\begin{equation}
f_{idiom} = \underset{i\in \{1,2,...,n_{w}\}}{\rm argTop}K_{idiom}\{freq(Re(w_{i},D_{idiom}))\},
\end{equation}
where $D_{modal}$ and $D_{idiom}$ represent the modal particle database and idiom database, respectively, and $Re(\cdot)$ denotes the regular expression operation used for matching. The function $freq(\cdot)$ calculates the frequency of each modal particle or idiom within the text. Identifying the top $K_{modal}$ modal particles and the top $K_{idiom}$ idioms based on their frequencies, we determine the predominant modal particles $f_{modal}$ and idioms $f_{idiom}$ that encapsulate the stylistic features of the text. This refined analysis not only quantifies the occurrence of modal particles and idioms but also provides a nuanced understanding of the author's linguistic choices and stylistic nuances.

After deriving various sub-features related to word characteristics, we integrate these features to construct a comprehensive representation of word attributes. This integrated approach allows for a holistic understanding of the textual style and linguistic nuances. To encapsulate the multi-faceted characteristics of words, we synthesize the previously computed sub-features into a unified feature representation. These features encompass part-of-speech, word length, syllabic words, modal particles, and idioms. This integration process is expressed as: 
\begin{equation}
f_{W} = \mathrm{Concat} \left( f_{ps} \oplus f_{l} \oplus f_{mo} \oplus f_{po} \oplus f_{modal} \oplus f_{idiom} \right),
\end{equation}
where $\oplus$ denotes the concatenation operator. This detailed concatenation ensures that all relevant sub-features are comprehensively integrated into the final feature $f_{W}$.

\subsubsection{\textbf{Sentence Level}}

Compared to fine-grained analysis at the word level, a comprehensive examination of sentence style provides a more holistic understanding of the author's unique personality and characteristics in language expression \cite{Kakoma2020A, Gajda2022The}. This approach contributes to a clearer perception of the overall style of the text, offering a broader perspective for a profound understanding of the article's stylistic nuances. Our analysis focuses on sentence style from the perspectives of sentence length, emotion, sentence structure, and rhetoric.

\textbf{Sentences Length.} To evaluate the average sentence length, we define sentence terminators as periods, question marks, and exclamation marks, using regular expressions to segment the text into a set of sentences $S=(s_{1},s_{2},...,s_{n_{s}})$, where $n_{s}$ denotes the number of sentences. Furthermore, we utilize regular expressions to match Chinese and English characters, yielding a valid character set $C$. By computing the lengths of the sentence set and the character set, the average sentence length of the article can be determined. This is formalized as:
\begin{equation}
f_{len} = len(C)/len(S),
\end{equation}
where $len(\cdot)$ denotes the length of the respective sets.

Furthermore, to analyze the distribution of sentence lengths throughout the article, we classify each sentence as either short or long based on on a threshold parameter $\theta$, which is set to 20 words following prior studies \cite{Karya2019A, Wallwork2016}. This classification enables us to capture the tone conveyed by the sentence lengths. The classification process is represented as follows:
\begin{equation}
f_{ls} = \underset{i\in \{1,2,...,n_{s}\}}{\rm argmax}\{I_{ls}(s_{i})\},
\end{equation}
where $I_{ls}(\cdot)$ is a function that counts the occurrences of long and short sentences. The function $I_{ls}(s_{i})$ is defined as:
\begin{equation}
I_{ls}(s_{i}) = \sum_{j=1}^{n_{s}} \delta(\mathrm{len}(s_{j}) - \theta),
\end{equation}
where $\delta(\mathrm{len}(s_{j}) - \theta)$ is an indicator function that returns 1 if the length of sentence $s_{j}$ exceeds $\theta$ words (classifying it as a long sentence), and 0 otherwise. If the article predominantly contains long sentences, it is classified as having a long sentence tone; conversely, if short sentences are more frequent, it is classified as having a short sentence tone. If the numbers are roughly equal, the article is considered to have a balanced combination of both long and short sentences.

\textbf{Emotion.} To analyze the emotional tone conveyed by the sentences, we utilize the ERNIE\footnote{https://github.com/PaddlePaddle/ERNIE} pre-trained model, fine-tuned on the OCEMOTION dataset\footnote{https://aistudio.baidu.com/datasetdetail/100731}. This model provides a more granular understanding of text semantics and emotional nuances by classifying emotions into seven fine-grained categories: \textit{sadness}, \textit{happiness}, \textit{disgust}, \textit{anger}, \textit{like}, \textit{surprise}, and \textit{fear}. To improve the granularity and accuracy of emotion detection, we updated ERNIE’s weights during fine-tuning on OCEMOTION. By aggregating the scores for each sentence, we determine the predominant emotion of the entire text. This process is formalized as follows:
\begin{equation}
f_{em} = \underset{j \in \{1, 2, \ldots, 7\}}{\mathrm{argmax}} \left\{ \sum_{i=1}^{n_{s}} I_{em}(\text{Emotion}_{\text{M}}(s_{i}, e_j)) \right\},
\end{equation}
where $\text{Emotion}_{\text{M}}(s_{i}, e_j)$ denotes the score of sentence $s_{i}$ for the $j$-th emotion, $\text{M}$ represent the $\text{ERNIE}$ model,$I_{em}(\cdot)$ is a function that calculates the total score for each emotion category $e_j$. The summation $\sum_{i=1}^{n_{s}} I_{em}(\text{Emotion}_{\text{M}}(s_{i}, e_j))$ aggregates the scores of all sentences for each emotion. The emotion with the highest total score is identified as the dominant emotion of the text. By leveraging this comprehensive emotional analysis, we can gain deeper insights into the emotional tone and expressive nuances of the text, providing a robust foundation for understanding the author's stylistic tendencies.

\textbf{Sentences Structure and Rhetoric.} 
To evaluate the structure of integrated and scattered sentences, we employ a deep learning model to make classification decisions. Utilizing the powerful text generation capabilities of ChatGPT, we design prompts to generate a classification dataset containing integrated and scattered sentences. This dataset is then used to fine-tune the MacBERT model\footnote{https://github.com/ymcui/MacBERT}, resulting in a specialized Chinese sentence structure binary classification model $BERT_{st}$. Each sentence in the article is classified to assess the distribution of integrated and scattered sentences. If the number of integrated sentences exceeds that of scattered sentences, it can be inferred that the article predominantly consists of integrated sentences, otherwise it comprises more scattered sentences. This is formalized as:
\begin{equation}
f_{st} = \underset{i\in \{1,2,...,n_{s}\}}{\rm argmax}\{I_{st}(
BERT_{st}(s_{i}))\},
\end{equation}
where $I_{st}(\cdot)$ is a function that calculates the counts of integrated and scattered sentences.

Similarly, we classify the rhetorical devices used in sentences by generating sentences with rhetorical devices such as metaphor, personification, exaggeration, citation, parallelism, irony, rhetorical questioning, statement, and duality using ChatGPT to form a fine-tuning dataset. This leads to the development of the Chinese rhetorical multi-classification model $BERT_{rhe}$. The classification of rhetorical devices in each sentence is given by:
\begin{equation}
f_{rhe} = \underset{i\in \{1,2,...,n_{s}\}}{\rm argTop}K_{rhe}\{I_{rhe}(
BERT_{rhe}(s_{i}))\},
\end{equation}
where $I_{rhe}(\cdot)$ is utilized for the statistical counting of various types of rhetorical sentences. The top $K_{rhe}$ rhetorical devices with the highest frequencies are identified as the main rhetorical features of the article. By leveraging these advanced models, we provide a detailed analysis of sentence structure and rhetorical usage, offering a comprehensive understanding of the text's stylistic attributes and enhancing the depth of linguistic analysis.

Subsequently, we integrate all sentence-level features to construct a comprehensive representation of sentence attributes. This integration is formalized as:
\begin{equation}
f_{S} = \mathrm{Concat} \left( f_{len} \oplus f_{ls} \oplus f_{em} \oplus f_{st} \oplus f_{rhe} \right),
\end{equation}
where $\oplus$ denotes the concatenation operator. Finally, we develop a holistic definition of the article's style by combining the features from both word and sentence levels. This comprehensive feature representation encapsulates the intricate stylistic elements of the text, enhancing our understanding of the author's unique linguistic expression. The final feature vector is constructed as:
\begin{equation}
F = \mathrm{Concat} \left( f_{W} \oplus f_{S} \right),
\end{equation}
where $f_{W}$ represents the word-level features, $f_{S}$ represents the sentence-level features.

\subsection{Style-enhanced Prompt}

At this stage, leveraging the article-style definition $F$ generated by the Text Style Definition (TSD) module, we design a style-enhanced prompt to augment the zero-shot learning capability of large language models (LLMs) for Chinese article-style transfer.

\begin{figure}[t]
  \centering
  \includegraphics[width=12cm]{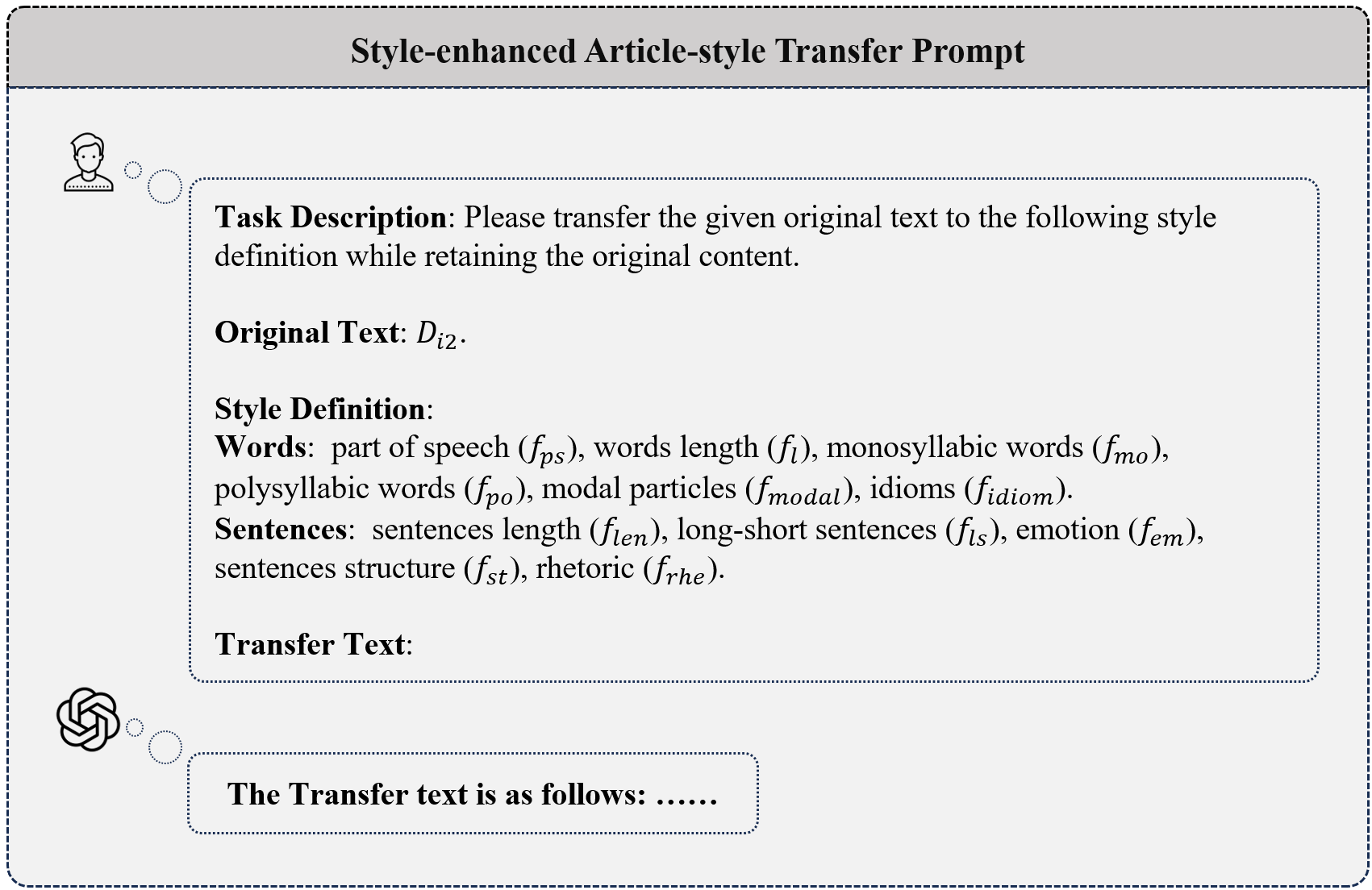}
   \caption{The style-enhanced article-style transfer prompt.}
   \label{fig3}
\end{figure}

The style-enhanced prompt comprises the task description, original text, style definition, and output indication. The complete format of the prompt is illustrated in Figure \ref{fig3}. Given that our text feature calculations primarily involve statistical analysis of vocabulary and syntactic structures, a mechanical enumeration of these structures may lead to misinterpretations by LLMs, resulting in unnecessary hallucinations. Therefore, we incorporate professional descriptions of feature structures to enable LLMs to better comprehend stylistic features. For instance, we might describe a style as containing many long sentences, which enrich the article with connotation, specificity in narrative, detailed reasoning, and emotional depth. For more specific details, please refer to the Appendix Figures \ref{figA1}-\ref{figA10}. Furthermore, considering the challenges LLMs face in understanding negative prompts, we meticulously design the task description to ensure that the generated text achieves an optimal balance between high transfer accuracy and content preservation, while avoiding the introduction of extraneous textual information. This approach helps maintain the integrity and authenticity of the original text, ensuring that the stylistic transfer is both accurate and contextually appropriate.

\section{Experiment}
\subsection{Datasets}

In our study, we carefully created diverse datasets consisting of texts from both literary and non-literary domains to ensure comprehensive validation of our framework.

For the literary works, we selected five seminal texts with distinct styles covering modern and contemporary China:
\begin{itemize}
    \item \textbf{Fortress Besieged}, authored by Qian Zhongshu, is a modern novel characterized by humor and satire.
    \item \textbf{The Scream}, written by Lu Xun, exemplifies realism in modern Chinese literature.
    \item \textbf{The Fifteen Year of the Wanli Era}, by Huang Renyu, is a historical narrative.
    \item \textbf{The Family Instructions of Zeng Guofan}, crafted by Zeng Guofan, represents classical family instructions within ancient literature.
    \item \textbf{The Three-Body Problem}, authored by Liu Cixin, is a renowned science fiction novel.
\end{itemize}

To broaden the representativeness and diversity of our dataset, we incorporated five additional datasets from various non-literary domains:
\begin{itemize}
    \item \textbf{News Dataset}: Sourced from \textit{news2016zh}\footnotemark[8], this dataset includes modern Chinese news articles on topics such as politics, technology, and culture, representing formal and concise journalistic writing.
    \item \textbf{Encyclopedic Dataset}: The Chinese Wikipedia\footnotemark[8] corpus provides factual and objective content written in a formal, highly structured style typical of encyclopedic entries.
    \item \textbf{Social Media Dataset}: Sourced from Baidu's AI Studio platform\footnotemark[9], this dataset includes user-generated posts from Chinese social media platforms, reflecting informal, conversational, and often expressive styles.
    \item \textbf{Legal Documents Dataset}: The \textit{CAIL2018}\footnotemark[10] dataset contains Chinese legal documents, such as court rulings, exemplifying technical and domain-specific writing.
    \item \textbf{Scientific Literature Dataset}: The \textit{CSL3}\footnotemark[10] dataset includes abstracts and articles from scientific journals, showcasing a formal and technical writing style suited for academic communication.
\end{itemize}

\footnotetext[8]{https://github.com/brightmart/nlp\_chinese\_corpus}
\footnotetext[9]{https://aistudio.baidu.com/datasetdetail/140467}
\footnotetext[10]{https://github.com/OYE93/Chinese-NLP-Corpus}

We divide each type of text into two main parts: the style transfer part and the style definition part. We create ten parallel Chinese article-style transfer datasets by transferring the style definition part into styleless text using ChatGPT. The prompt used for ChatGPT is shown in Appendix Figure \ref{figA11}. The number of test samples for each book is approximately 400, with each sample consisting of 120 to 140 words. Table \ref{tab1} presents detailed statistics of the datasets.

% Table generated by Excel2LaTeX from sheet 'Dataset'
\begin{table}[t]
  \centering
  \caption{Statistical overview of ten different datasets.}
  \resizebox{\textwidth}{!}{
  \renewcommand{\arraystretch}{1.2}
    \begin{tabular}{ccc}
    \toprule
    Chinese texts & Numbers of style transfer examples & Numbers of style definition examples \\
    \midrule
    \midrule
    Fortress Besieged & 623   & 946 \\
    The Scream & 217   & 284 \\
    The Fifteen Year of the Wanli Era & 413   & 617 \\
    The Family Instructions of Zeng Guofan & 302   & 1026 \\
    The Three-Body Problem & 513   & 1762 \\
    News  & 413   & 516 \\
    Wikipedia & 221   & 112 \\
    Social Media & 330   & 324 \\
    Legal Document & 439   & 455 \\
    Scientific Literature & 401   & 407 \\
    \bottomrule
    \end{tabular}}
  \label{tab1}%
\end{table}%

\subsection{Implementation Details}

For the algorithm parameters at the word level, we set $K_{l}$, $K_{sy}$, $K_{modal}$, $K_{idiom}$ to 4, 2, 6, 8, respectively. At the sentence level, we designate $K_{rhe}$ as 2. The integrated and scattered sentences dataset we collected comprises 600 instances per class for binary classification, while the rhetoric dataset consists of 600 instances per class for ten-class classification, encompassing nine rhetorical devices and a class without any rhetorical devices. We use these two datasets to fine tune BERT separately on four RTX 3090 GPUs. In the composition of text style definition $F$, the stylistic definition at the word level $f_{W}$ is positioned before the sentences level style definition $f_{S}$. For the configuration of LLMs, the temperature parameter is set to 0.5, and the seed parameter is uniformly set to 42.

\subsection{Evaluation Metrics}

Our evaluation primarily focus on \textbf{style transfer accuracy} and \textbf{content preservation}. A high-quality Chinese article-style transfer model should balance these two aspects. We compare the generated text with the original text of the article to calculate the transfer accuracy. Simultaneously, comparisons between the generated text and the styleless text are conducted to assess the degree of content preservation. The evaluation metrics adopted for both perspectives are exactly the same. Specifically, we measure the lexical and semantic similarity between two texts using Bilingual Evaluation Understudy (BLEU)-n (n=1,2) \cite{papineni2002} and BERTScore \cite{bert-score}, where BERTScore mainly includes Precision, Recall, and F1 score. The specific calculation process is as follows:

\textbf{BLEU}: BLEU is a metric used to evaluate the quality of text by measuring the n-gram overlap between the generated text and reference texts. It is widely used for assessing the performance of machine translation and other text generation tasks, including style transfer. The general formula for BLEU is:
\begin{equation}
\text{BLEU} = BP \cdot \exp \left( \sum_{n=1}^{N} w_n \log p_n \right),
\end{equation}
where $BP$ (Brevity Penalty) is a factor that penalizes overly short translations. $w_n$ is the weight for the n-gram precision, typically $w_{n} = 1/N$ when uniform weights are used. $p_n$ is the precision for n-grams.

\textbf{BERTScore}: BERTScore leverages BERT embeddings to evaluate the similarity between generated text and reference text, capturing both lexical and semantic nuances. This metric is particularly suited for style transfer tasks where semantic fidelity is critical. The formula for BERTScore is: 
\begin{equation}
\text{BERTScore} = \frac{1}{|T|} \sum_{t \in T} \max_{r \in R} \{\text{Sim}(\text{BERT}(t), \text{BERT}(r))\},
\end{equation}
where $T$ is the set of tokens in the generated text, $R$ is the set of tokens in the reference text, $\text{BERT}(t)$ represents the BERT embedding of token $t$, $\text{Sim}(a,b)$ is the cosine similarity between vectors $a$ and $b$, defined as:
\begin{equation}
\text{Sim}(a, b) = \frac{a \cdot b}{\|a\| \|b\|}.
\end{equation}

In addition, considering the randomness of LLMs generating text, we transfer each type of text ten times to calculate the final average metrics results.

\subsection{Baselines}
Considering that our algorithm primarily targets Chinese text, we have selected several state-of-the-art LLMs for generating Chinese text, including the GPT series, ChatGLM series, and Baichuan series. Detailed information on LLMs is shown in Table \ref{tab2}. The GPT series, developed by OpenAI, consists of a series of close LLMs, primarily including GPT-3.5 and GPT-4. These models are among the most widely used in the field. In this study, we utilized \textbf{GPT-4-1106-preview} and \textbf{GPT-3.5-turbo} as key components of our LLMs framework. The ChatGLM series of models were jointly developed by Tsinghua University and Zhichart AI Company. In this research, we utilized the \textbf{ChatGLM3-6B-chat} model to assist in generating Chinese text. The Baichuan series encompasses a range of advanced multilingual base language models developed by ``Baichuan Intelligence''. In this study, we employed the open-source \textbf{Baichuan-13B-chat} model.

In addition, to comprehensively evaluate our proposed framework, we conduct a comparative study on six state-of-the-art baseline models. \textbf{Style Transformer} \cite{dai-etal} employs a transformer-based architecture to disentangle content and style, while \textbf{RACoLN} \cite{Dongkyu-etal} utilizes reinforcement learning to balance style accuracy and content preservation. \textbf{StyleLM} \cite{syed2020adapting} leverages a pre-trained encoder-decoder model fine-tuned with a denoising autoencoder loss to enable stylized text rewriting without parallel data. \textbf{SDR} \cite{zheng2021stylized} generates pseudo-dialogue pairs to train a stylized dialogue model for coherent, style-aligned responses. \textbf{DRAG} \cite{singh2021drag} introduces a director-generator framework to propose and refine style-aligned text variants, and \textbf{StoryTrans} \cite{zhu2023storytrans} disentangles stylistic features from discourse representations, using a mask-and-fill framework to enhance content preservation. Through comparisons between the proposed CAT-LLM framework and these diverse state-of-the-art baselines, we can more accurately assess the strengths and weaknesses of CAT-LLM.

\begin{table}[t]
  \centering
  \renewcommand\arraystretch{1.1}
  \caption{Specific parameter details for Chinese LLMs.}
  \begin{threeparttable}
    \begin{tabular}{ccccc}
    \toprule
    \textbf{Chinese LLMs} & \textbf{Parm.} & \textbf{Type} & \textbf{Publisher} & \textbf{Release} \\
    \midrule
    \midrule
    Baichuan2 & 13B   & Chat  & Baichuan Inc. & 2023.09 \\
    ChatGLM3 & 6B    & Chat  & Tsinghua & 2023.10 \\
    GPT-3.5-Turbo & 175B  & Chat  & OpenAI & 2023.03 \\
    GPT-4-1106 & NaN   & Chat  & OpenAI & 2023.11 \\
    \bottomrule
    \end{tabular}
    \begin{tablenotes}
        \footnotesize
        \item \textit{Note}: NaN denotes no public data available. 
    \end{tablenotes}
  \end{threeparttable}
  \label{tab2}%
\end{table}%

\subsection{Automatic Evaluation Results}

\begin{figure}[t]
\centering
\subfigure[Social Media.]{ 
\label{fig4.1}
\includegraphics[width=0.49\textwidth]{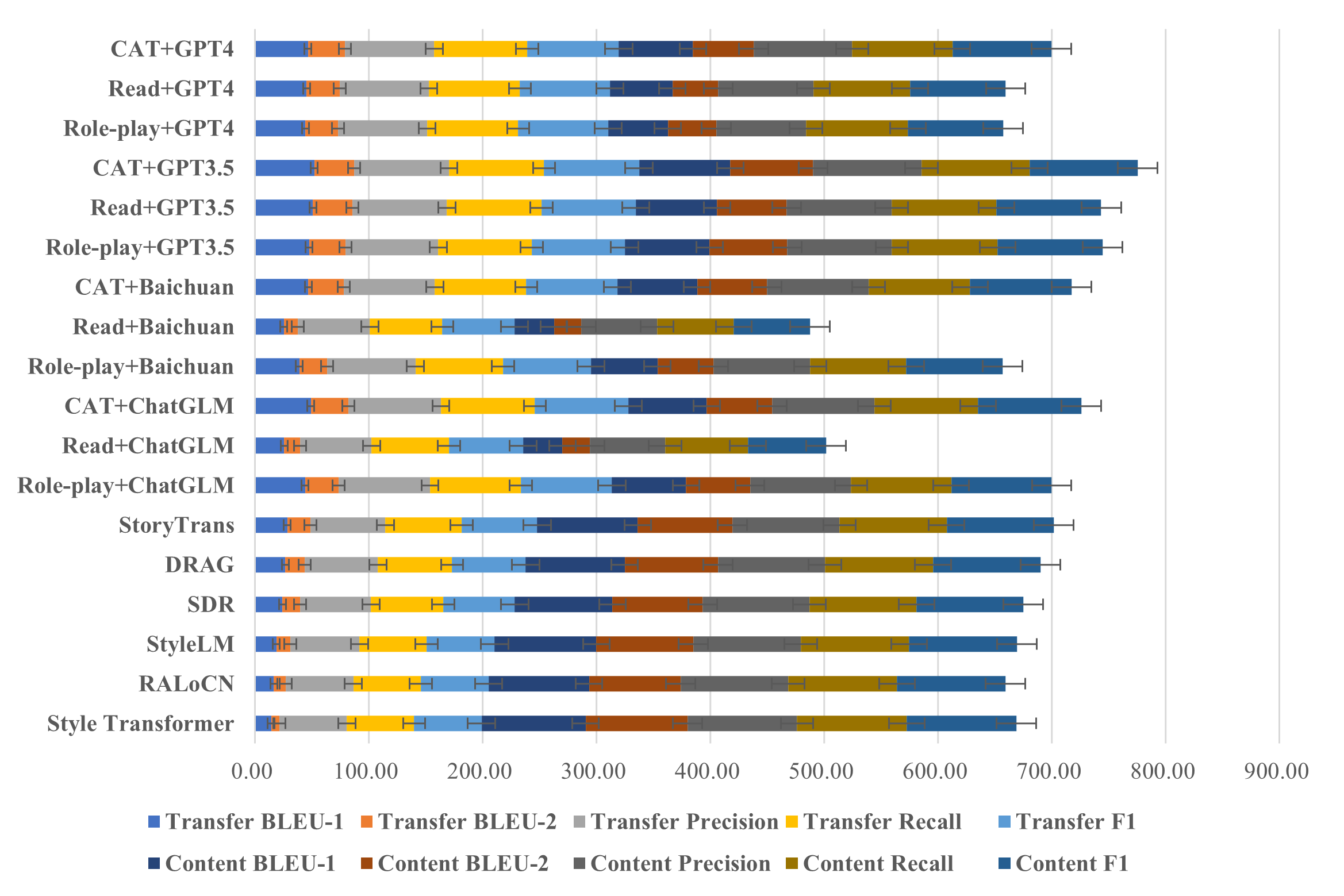}}
\subfigure[Legal Document.]{
\label{fig4.2}
\includegraphics[width=0.49\textwidth]{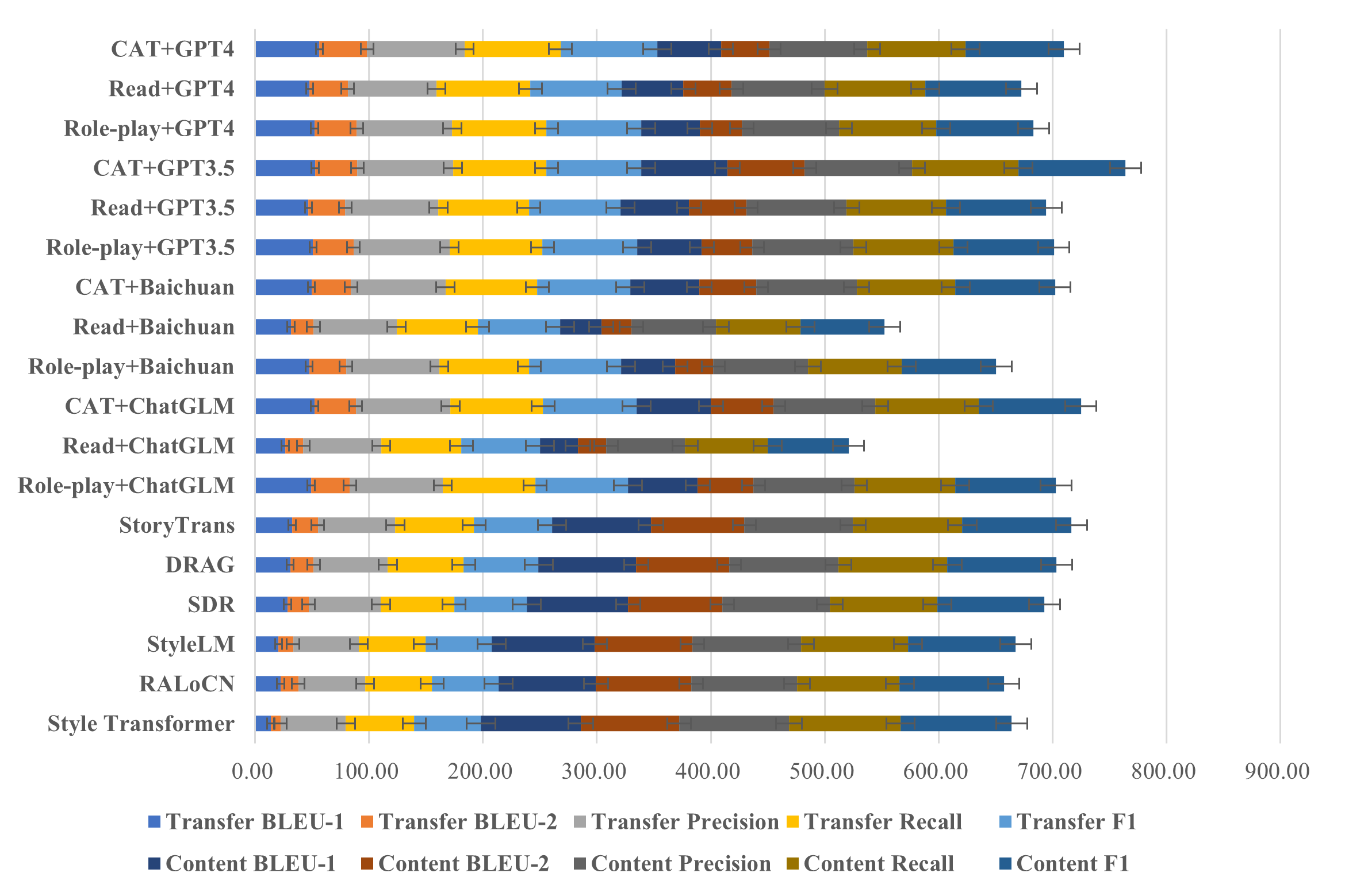}}
\caption{Stacked bar chart representation of various models' performance on ``Social Media'' and ``Legal Document'' datasets.}
\label{fig4}
\end{figure}

Our experimental section aims to evaluate the performance of different style transfer models, including the proposed CAT-LLM, models based on LLMs directly reading articles, role-playing under different LLMs, and six state-of-the-art baselines. The objective of these experiments is to explore how each model balances style transfer and content preservation and their ability to precisely control style transfer in generated text. To ensure the fairness of the experiments, we maintain consistency in the tokens used for reading articles in experiments involving LLMs directly reading the articles with the style definition tokens of CAT-LLM. This ensures a fair and reasonable comparison between different models. As shown in Table \ref{tab3}, we present the experimental results of the models on three selected datasets: \textit{Fortress Besieged}, \textit{News}, and \textit{Wikipedia}, chosen from the ten datasets. In addition, the experimental results of the LLMs on the selected \textit{Social Media} and \textit{Legal Document} datasets are presented in stacked bar charts shown in Figure \ref{fig4}, which provide a clear overview of the contributions of different metrics to the overall performance, as well as the overall performance of each model. The test results of all models on other datasets are detailed in Appendix Tables \ref{tabA1}-\ref{tabA5}. CAT-LLM, based on various LLMs, achieves a better balance between style transfer and content preservation in each LLM transfer, providing more precise control over style transfer in generated text. We used multiple datasets to evaluate the performance of the models, covering different styles and content to ensure the generalizability of the results. Evaluation metrics include the accuracy of style transfer, and the degree of content preservation.

% Table generated by Excel2LaTeX from sheet 'Sheet3'
\begin{table}[t]
  \centering
  \caption{Automatic evaluation results on the ``Fortress Besieged'', ``News'', and ``Wikipedia'' datasets.}
  \renewcommand\arraystretch{1.1}
  \resizebox{\textwidth}{!}{
    \begin{tabular}{ccccccccccccc}
    \toprule
    \multirow{2}[4]{*}{\textbf{Chinese Text}} & \multirow{2}[4]{*}{\textbf{Models}} & \multicolumn{5}{c}{\textbf{Transfer Accuracy(\%)}} &       & \multicolumn{5}{c}{\textbf{Content Preservation(\%)}} \\
\cmidrule{3-7}\cmidrule{9-13}          &       & \textbf{BLEU-1} & \textbf{BLEU-2} & \textbf{Precision} & \textbf{Recall} & \textbf{F1} &       & \textbf{BLEU-1} & \textbf{BLEU-2} & \textbf{Precision} & \textbf{Recall} & \textbf{F1} \\
    \midrule
    \midrule
    \multicolumn{1}{c}{\multirow{18}[4]{*}{\textbf{Fortress Besieged}}} & Style Transformer & 17.80  & 11.03  & 55.45  & 58.79  & 57.08  &       & \textbf{95.34} & \textbf{90.93} & \textbf{96.87} & \textbf{96.50} & \textbf{96.69} \\
          & RALoCN & 22.86  & 17.01  & 58.52  & 58.71  & 58.70  &       & 92.27  & 88.07  & 95.34  & 95.84  & 95.59  \\
          & StyleLM & 20.54  & 16.91  & 56.40  & 57.09  & 56.78  &       & 93.29  & 88.43  & 95.12  & 95.38  & 95.25  \\
          & SDR   & 30.11  & 21.68  & 60.92  & 62.17  & 61.66  &       & 91.31  & 80.79  & 93.02  & 92.27  & 92.74  \\
          & DRAG  & 33.20  & 22.45  & 63.95  & 64.92  & 64.46  &       & 91.18  & 86.33  & 94.92  & 95.20  & 95.03  \\
          & StoryTrans & \textbf{34.17} & \textbf{23.55} & \textbf{67.82} & \textbf{68.01} & \textbf{67.93} &       & 90.02  & 84.37  & 94.73  & 94.98  & 94.86  \\
\cmidrule{2-7}\cmidrule{9-13}          & Role-play+ChatGLM & 33.37  & 18.36  & 71.18  & 72.26  & 71.68  &       & 56.85  & 50.51  & 81.58  & 82.70  & 82.10  \\
          & Read+ChatGLM & 21.68  & 7.36  & 60.96  & 62.66  & 61.77  &       & 28.01  & 14.39  & 63.72  & 64.69  & 64.15  \\
          & \cellcolor{Ocean}{\textbf{CAT+ChatGLM}} & \cellcolor{Ocean}{43.81}  & \cellcolor{Ocean}{25.40}  & \cellcolor{Ocean}{77.06}  & \cellcolor{Ocean}{78.93}  & \cellcolor{Ocean}{77.97}  &   \cellcolor{Ocean}{}    & \cellcolor{Ocean}{82.80}  & \cellcolor{Ocean}{77.15}  & \cellcolor{Ocean}{94.33}  & \cellcolor{Ocean}{94.37}  & \cellcolor{Ocean}{94.34}  \\
          & Role-play+Baichuan & 35.13  & 16.54  & 72.18  & 73.88  & 73.01  &       & 85.86  & 82.24  & 87.45  & 87.15  & 87.29  \\
          & Read+Baichuan & 17.71  & 6.91  & 60.23  & 63.91  & 62.00  &       & 29.91  & 21.92  & 64.27  & 67.34  & 65.74  \\
          & \cellcolor{Ocean}{\textbf{CAT+Baichuan}} & \cellcolor{Ocean}{42.61}  & \cellcolor{Ocean}{25.07}  & \cellcolor{Ocean}{76.11}  & \cellcolor{Ocean}{77.17}  & \cellcolor{Ocean}{76.59}  &   \cellcolor{Ocean}{}    & \cellcolor{Ocean}{83.58}  & \cellcolor{Ocean}{80.13}  & \cellcolor{Ocean}{93.59}  & \cellcolor{Ocean}{92.71}  & \cellcolor{Ocean}{93.11}  \\
          & Role-play+GPT3.5 & 38.57  & 20.67  & 74.81  & 76.70  & 75.73  &       & 64.40  & 53.79  & 88.32  & 88.63  & 88.46  \\
          & Read+GPT3.5 & 16.59  & 6.76  & 59.77  & 63.87  & 61.74  &       & 27.03  & 19.05  & 64.21  & 68.09  & 66.07  \\
          & \cellcolor{Ocean}{\textbf{CAT+GPT3.5}} & \cellcolor{Ocean}{\textbf{47.63}} & \cellcolor{Ocean}{\textbf{27.05}} & \cellcolor{Ocean}{\textbf{78.74}} & \cellcolor{Ocean}{\textbf{79.93}} & \cellcolor{Ocean}{\textbf{79.36}} &  \cellcolor{Ocean}{}     & \cellcolor{Ocean}{\textbf{88.96}} & \cellcolor{Ocean}{\textbf{85.16}} & \cellcolor{Ocean}{\textbf{96.24}} & \cellcolor{Ocean}{\textbf{96.62}} & \cellcolor{Ocean}{\textbf{96.47}} \\
          & Role-play+GPT4 & 27.10  & 10.85  & 69.40  & 71.64  & 70.48  &       & 32.39  & 16.73  & 74.57  & 75.82  & 75.17  \\
          & Read+GPT4 & 28.85  & 13.40  & 69.38  & 73.15  & 71.14  &       & 43.68  & 28.45  & 74.87  & 79.15  & 76.91  \\
          & \cellcolor{Ocean}{\textbf{CAT+GPT4}} & \cellcolor{Ocean}{35.12}  & \cellcolor{Ocean}{17.23}  & \cellcolor{Ocean}{70.68}  & \cellcolor{Ocean}{73.22}  & \cellcolor{Ocean}{72.08}  &  \cellcolor{Ocean}{}     & \cellcolor{Ocean}{50.09}  & \cellcolor{Ocean}{34.62}  & \cellcolor{Ocean}{82.47}  & \cellcolor{Ocean}{82.95}  & \cellcolor{Ocean}{82.67}  \\
    \midrule
    \midrule
    \multicolumn{1}{c}{\multirow{18}[4]{*}{\textbf{News}}} & Style Transformer & 11.82  & 7.47  & 50.06  & 51.71  & 50.87  &       & \textbf{90.93} & \textbf{87.47} & 92.06  & 88.88  & 91.07  \\
          & RALoCN & 12.34  & 11.55  & 58.75  & 55.83  & 57.45  &       & 89.67 & 85.77  & 88.36  & 80.97  & 85.02  \\
          & StyleLM & 17.75  & 10.08  & 61.31  & 62.32  & 61.87  &       & 87.45 & 82.15  & 92.03  & 86.69  & 89.09  \\
          & SDR   & 21.80  & 14.83  & 67.82  & 66.69  & 65.74  &       & 84.00  & 81.28  & 92.45  & 86.86  & 89.70  \\
          & DRAG  & 24.93  & 21.14  & 71.26  & 73.90  & 72.47  &       & 79.86  & 68.76  & \textbf{96.51} & 89.41  & 93.66  \\
          & StoryTrans & \textbf{29.48} & \textbf{22.48} & \textbf{75.41} & \textbf{78.11} & \textbf{77.07} &       & 81.94  & 68.32  & 96.11  & \textbf{94.43} & \textbf{95.27} \\
\cmidrule{2-7}\cmidrule{9-13}          & Role-play+ChatGLM & 50.88  & 30.49  & 83.06  & 80.15  & 82.08  &       & 61.41  & 48.18  & 88.73  & 86.16  & 87.43  \\
          & Read+ChatGLM & 31.32  & 20.04  & 68.68  & 76.58  & 72.35  &       & 38.40  & 31.47  & 71.64  & 83.38  & 76.98  \\
          & \cellcolor{Ocean}{\textbf{CAT+ChatGLM}} & \cellcolor{Ocean}{54.66}  & \cellcolor{Ocean}{37.27}  & \cellcolor{Ocean}{83.23}  & \cellcolor{Ocean}{82.31}  & \cellcolor{Ocean}{82.75}  &  \cellcolor{Ocean}{}     & \cellcolor{Ocean}{71.83}  & \cellcolor{Ocean}{61.34}  & \cellcolor{Ocean}{90.59}  & \cellcolor{Ocean}{91.25}  & \cellcolor{Ocean}{90.89}  \\
          & Role-play+Baichuan & 38.37  & 24.13  & 79.30  & 73.68  & 76.14  &       & 43.06  & 30.25  & 81.81  & 77.19  & 79.16  \\
          & Read+Baichuan & 26.39  & 9.38  & 63.70  & 64.03  & 63.86  &       & 29.41  & 14.31  & 64.85  & 66.11  & 65.46  \\
          & \cellcolor{Ocean}{\textbf{CAT+Baichuan}} & \cellcolor{Ocean}{53.47}  & \cellcolor{Ocean}{35.96}  & \cellcolor{Ocean}{82.86}  & \cellcolor{Ocean}{81.37}  & \cellcolor{Ocean}{82.08}  &   \cellcolor{Ocean}{}    & \cellcolor{Ocean}{72.11}  & \cellcolor{Ocean}{62.15}  & \cellcolor{Ocean}{90.36}  & \cellcolor{Ocean}{90.26}  & \cellcolor{Ocean}{90.29}  \\
          & Role-play+GPT3.5 & 51.49  & 34.73  & 83.39  & 80.93  & 82.11  &       & 62.08  & 49.14  & 88.80  & 87.55  & 88.14  \\
          & Read+GPT3.5 & 53.55  & 36.30  & 82.91  & 81.38  & 82.12  &       & 70.69  & 60.34  & 90.41  & 90.25  & 90.32  \\
          & \cellcolor{Ocean}{\textbf{CAT+GPT3.5}} & \cellcolor{Ocean}{\textbf{57.91}} & \cellcolor{Ocean}{\textbf{40.41}} & \cellcolor{Ocean}{\textbf{85.14}} & \cellcolor{Ocean}{\textbf{83.83}} & \cellcolor{Ocean}{\textbf{84.47}} &   \cellcolor{Ocean}{}    & \cellcolor{Ocean}{\textbf{83.16}} & \cellcolor{Ocean}{\textbf{76.54}} & \cellcolor{Ocean}{\textbf{95.28}} & \cellcolor{Ocean}{\textbf{95.47}} & \cellcolor{Ocean}{\textbf{95.37}} \\
          & Role-play+GPT4 & 50.30  & 32.49  & 81.09  & 79.96  & 80.49  &       & 54.90  & 38.71  & 83.67  & 83.95  & 83.78  \\
          & Read+GPT4 & 50.59  & 33.85  & 78.49  & 82.48  & 80.41  &       & 60.63  & 45.52  & 84.18  & 86.35  & 85.24  \\
          & \cellcolor{Ocean}{\textbf{CAT+GPT4}} & \cellcolor{Ocean}{53.73}  & \cellcolor{Ocean}{35.53}  & \cellcolor{Ocean}{80.93}  & \cellcolor{Ocean}{81.44}  & \cellcolor{Ocean}{81.17}  &  \cellcolor{Ocean}{}     & \cellcolor{Ocean}{61.63}  & \cellcolor{Ocean}{50.43}  & \cellcolor{Ocean}{83.12}  & \cellcolor{Ocean}{89.52}  & \cellcolor{Ocean}{86.17}  \\
    \midrule
    \midrule
    \multicolumn{1}{c}{\multirow{18}[4]{*}{\textbf{Wikipedia}}} & Style Transformer & 10.41  & 9.75  & 58.93  & 59.12  & 59.03  &       & \textbf{85.43} & \textbf{83.29} & \textbf{98.19} & \textbf{98.09} & \textbf{98.14} \\
          & RALoCN & 18.39  & 11.72  & 62.65  & 62.74  & 62.69  &       & 82.03  & 80.32  & 95.08  & 95.81  & 95.45  \\
          & StyleLM & 21.77  & 13.46  & 68.01  & 67.29  & 67.66  &       & 85.06  & 80.92  & 93.88  & 95.18  & 94.79  \\
          & SDR   & 23.21  & 18.06  & 74.88  & 72.34  & 73.59  &       & 84.93  & 78.48  & 92.78  & 91.88  & 92.15  \\
          & DRAG  & 25.74  & 21.38  & 77.47  & 77.77  & 77.62  &       & 83.56  & 79.62  & 95.92  & 94.98  & 95.40  \\
          & StoryTrans & \textbf{28.01} & \textbf{24.49} & \textbf{80.92} & \textbf{81.46} & \textbf{81.21} &       & 82.35  & 77.62  & 94.55  & 93.49  & 93.92  \\
\cmidrule{2-7}\cmidrule{9-13}          & Role-play+ChatGLM & 50.36  & 37.06  & 83.52  & 82.65  & 83.06  &       & 66.51  & 53.38  & 89.66  & 88.84  & 89.23  \\
          & Read+ChatGLM & 36.66  & 23.71  & 69.63  & 75.99  & 72.61  &       & 51.52  & 45.33  & 74.88  & 83.82  & 79.03  \\
          & \cellcolor{Ocean}{\textbf{CAT+ChatGLM}} & \cellcolor{Ocean}{52.93}  & \cellcolor{Ocean}{35.70}  & \cellcolor{Ocean}{83.05}  & \cellcolor{Ocean}{82.24}  & \cellcolor{Ocean}{82.63}  &  \cellcolor{Ocean}{}     & \cellcolor{Ocean}{68.17}  & \cellcolor{Ocean}{55.89}  & \cellcolor{Ocean}{90.52}  & \cellcolor{Ocean}{89.69}  & \cellcolor{Ocean}{90.09}  \\
          & Role-play+Baichuan & 51.12  & 33.34  & 81.52  & 81.57  & 81.51  &       & 56.96  & 41.15  & 84.78  & 85.00  & 84.86  \\
          & Read+Baichuan & 37.98  & 22.57  & 73.23  & 71.85  & 72.50  &       & 43.97  & 30.61  & 76.43  & 75.16  & 75.76  \\
          & \cellcolor{Ocean}{\textbf{CAT+Baichuan}} & \cellcolor{Ocean}{51.50}  & \cellcolor{Ocean}{34.42}  & \cellcolor{Ocean}{82.11}  & \cellcolor{Ocean}{81.28}  & \cellcolor{Ocean}{81.61}  &  \cellcolor{Ocean}{}     & \cellcolor{Ocean}{64.85}  & \cellcolor{Ocean}{52.71}  & \cellcolor{Ocean}{88.90}  & \cellcolor{Ocean}{87.95}  & \cellcolor{Ocean}{88.40}  \\
          & Role-play+GPT3.5 & 55.89  & 38.88  & 84.55  & 83.73  & 84.11  &       & 65.48  & 52.75  & 90.00  & 89.23  & 89.60  \\
          & Read+GPT3.5 & 52.51  & 35.83  & 82.90  & 81.53  & 82.19  &       & 67.06  & 56.40  & 90.29  & 88.84  & 89.53  \\
          & \cellcolor{Ocean}{\textbf{CAT+GPT3.5}} & \cellcolor{Ocean}{\textbf{56.94}} & \cellcolor{Ocean}{\textbf{39.24}} & \cellcolor{Ocean}{\textbf{85.84}} & \cellcolor{Ocean}{\textbf{84.95}} & \cellcolor{Ocean}{\textbf{85.43}} &  \cellcolor{Ocean}{}     & \cellcolor{Ocean}{\textbf{72.53}} & \cellcolor{Ocean}{\textbf{63.17}} & \cellcolor{Ocean}{\textbf{93.25}} & \cellcolor{Ocean}{\textbf{92.22}} & \cellcolor{Ocean}{\textbf{92.72}} \\
          & Role-play+GPT4 & 49.46  & 33.65  & 82.16  & 79.91  & 80.89  &       & 55.01  & 41.91  & 86.07  & 83.86  & 84.82  \\
          & Read+GPT4 & 42.32  & 27.29  & 74.79  & 81.63  & 77.99  &       & 48.23  & 35.60  & 78.68  & 86.64  & 82.39  \\
          & \cellcolor{Ocean}{\textbf{CAT+GPT4}} & \cellcolor{Ocean}{55.76}  & \cellcolor{Ocean}{37.98}  & \cellcolor{Ocean}{83.79}  & \cellcolor{Ocean}{83.84}  & \cellcolor{Ocean}{83.80}  &  \cellcolor{Ocean}{}     & \cellcolor{Ocean}{61.73}  & \cellcolor{Ocean}{46.76}  & \cellcolor{Ocean}{87.45}  & \cellcolor{Ocean}{87.64}  & \cellcolor{Ocean}{87.52}  \\
    \bottomrule
    \end{tabular}}
  \label{tab3}
\end{table}%

In terms of style transfer accuracy, the combined application of CAT+GPT-3.5-Turbo achieves the best performance on all datasets. In terms of lexical similarity evaluation index BLEU-1, CAT+GPT-3.5-Turbo has all reached about 50\%, while maintaining a level of around 80\% in semantic similarity. This success is not only attributed to the outstanding capabilities of GPT-3.5-Turbo but also highlights the effectiveness of the TSD module's style prompts, enabling CAT+GPT-3.5-Turbo to excel in article-style transfer. However, the experimental results of LLMs directly reading article fragments for style transfer remain unsatisfactory, with both style transfer accuracy and content preservation falling short of classic baselines such as Storytrans. This is mainly because the LLMs only summarize the style of certain fragments of the article and lack understanding of the overall style, illustrating the challenges LLMs face in handling more abstract and deeper semantic relationships, reasoning, and text comprehension. In terms of role-playing, LLMs excel at imitating writing styles of characters they are familiar with, while their performance significantly weakens when tasked with unfamiliar styles. As shown in Table \ref{tab3}, for news articles and Wikipedia texts, LLMs perform well in style transfer through role-playing. However, when asked to emulate the writing style of Qian Zhongshu's \textit{Fortress Besieged}, the results of style transfer are notably diminished. Furthermore, the transfer accuracy of the six baseline models is relatively low. While we acknowledge the parameter discrepancies between LLMs and traditional methods, it is important to recognize that LLMs represent the latest advancements in NLP, and comparing them with older techniques underscores the progress made in the field. In the era of LLMs, the integration of such models into future developments has become a prevailing trend. Thus, comparing the transfer performance of both types of models remains of practical significance. Upon analyzing the text generated by the baseline models, we observe that these models demonstrate a tendency to replicate the input text, selectively substituting certain words, but fail to effectively capture the semantic and stylistic nuances of the original content. This is the primary reason for their lower transfer accuracy, despite maintaining relatively higher content preservation.

Additionally, as shown in Table \ref{tab3} and Figure \ref{fig4}, GPT-4 does not perform as well as GPT-3.5-Turbo. For instance, in the style transfer of ``Fortress Besieged'', CAT+GPT-3.5-Turbo achieves a BLEU-1 score of 47.63\%, whereas CAT+GPT-4 only achieves 35.12\%. Similarly, in the ``News'' text, CAT+GPT-3.5-Turbo has a BERT-F1 score of 84.47\%, while CAT+GPT-4 achieves 81.17\%. This indicates that although GPT-4, as a more advanced model, is theoretically expected to perform better, its actual performance in style transfer tasks is not as good as anticipated, and it even falls short of GPT-3.5-Turbo. We speculate that this phenomenon may be due to several factors. Firstly, GPT-4, being a more advanced model, possesses higher intelligence and stronger language comprehension abilities. Therefore, during the generation process, it may adjust and optimize the input content to align with what it perceives as a more appropriate expression. This ``creative freedom'' enhances the readability and naturalness of the text but may lead to deviations from the required style definition in style transfer tasks, thereby affecting the accuracy of the transfer. Another possible reason lies in the differences in training data and optimization objectives. GPT-4 may have been exposed to more diverse and complex text data during training, with broader optimization goals that go beyond merely preserving and transferring text style. This results in GPT-4 focusing more on diversity and innovation in text generation in practical applications, at the expense of the precision required for style transfer tasks. Therefore, GPT-4's poor performance in style transfer tasks may be due to its higher language comprehension abilities and greater generation freedom. While these abilities are advantageous in many language tasks, they can become a disadvantage in tasks that require a high degree of style consistency.

With regard to content preservation, CAT-LLM has also achieved competitive results in the style transfer of various LLMs. In the experiment of directly reading articles with LLMs, we find the problem of introducing a large number of unrelated words, indicating that LLMs may be prone to hallucinations when dealing with complex problems. It is worth noting that the combination of our CAT-LLM also outperforms the Role-play+GPT-3.5-Turbo, indicating that the proposed TSD module can develop the enormous potential of LLMs in article-style transfer. The effectiveness of CAT-LLM in maintaining high content preservation is particularly evident across different datasets. For example, in the case of ``Wikipadia'' dataset, CAT+GPT-3.5-Turbo achieves a remarkable BERT-F1 score of 92.72\%, significantly higher than other models such as Role-play+GPT-3.5-Turbo, which scores 89.60\%. This indicates that while Role-play models focus on mimicking specific styles, they may sacrifice content accuracy, whereas CAT-LLM maintains a better balance. Moreover, the experiments reveal that smaller models such as Style Transformer and RACoLN, although showing higher content preservation scores, tend to merely replicate input texts with minimal stylistic changes. For instance, in ``Fortress Besieged'', the Style Transformer achieves a content preservation BERT-F1 score of 96.69\%, but its transfer accuracy remains low, suggesting that it prioritizes content retention over stylistic fidelity. This further highlights the superiority of CAT-LLM, which not only preserves content but also effectively transforms style, achieving a more holistic transfer. Additionally, the hallucination problem observed in direct LLM applications underscores the necessity for mechanisms like the TSD module. By providing structured prompts and guidance, the TSD module helps mitigate the risk of generating irrelevant content, thereby enhancing the reliability of style transfer tasks. This capability is crucial for practical applications where maintaining the integrity and coherence of the original content is as important as achieving the desired stylistic transformation.

% Table generated by Excel2LaTeX from sheet 'Sheet4'
\begin{table}[b]
  \centering
  \renewcommand\arraystretch{1.0}
  \caption{Human evaluation results on the ``Fortress Besieged'', ``News'', and ``Wikipedia'' datasets.}
  \resizebox{10cm}{!}{
    \begin{tabular}{ccccc}
    \toprule
    \textbf{Chinese Text} & \textbf{Models} & \textbf{Transfer Accuracy} & \textbf{Content Presevation} & \textbf{Fluency} \\
    \midrule
    \midrule
    \multicolumn{1}{c}{\multirow{18}[4]{*}{\textbf{Fortress Besieged}}} & Style Transformer & 40    & 87    & 65  \\
          & RALoCN & 45    & 85    & 67  \\
          & StyleLM & 50    & 84    & 68  \\
          & SDR   & 55    & 88    & 70  \\
          & DRAG  & 60    & 90    & 72  \\
          & StoryTrans & 70    & 91    & 74  \\
\cmidrule{2-5}          & Role-play+ChatGLM & 78    & 82    & 77  \\
          & Read+ChatGLM & 80    & 83    & 78  \\
          & \cellcolor{Ocean}{\textbf{CAT+ChatGLM}} & \cellcolor{Ocean}{85}    & \cellcolor{Ocean}{84}    & \cellcolor{Ocean}{80}  \\
          & Role-play+Baichuan & 75    & 81    & 75  \\
          & Read+Baichuan & 72    & 79    & 74  \\
          & \cellcolor{Ocean}{\textbf{CAT+Baichuan}} & \cellcolor{Ocean}{88}    & \cellcolor{Ocean}{83}    & \cellcolor{Ocean}{76}  \\
          & Role-play+GPT3.5 & 82    & 88    & 85  \\
          & Read+GPT3.5 & 85    & 90    & 85  \\
          & \cellcolor{Ocean}{\textbf{CAT+GPT3.5}} & \cellcolor{Ocean}{\textbf{93}} & \cellcolor{Ocean}{\textbf{96}} & \cellcolor{Ocean}{88}  \\
          & Role-play+GPT4 & 90    & 94    & 92  \\
          & Read+GPT4 & 92    & 95    & 93  \\
          & \cellcolor{Ocean}{\textbf{CAT+GPT4}} & \cellcolor{Ocean}{91}    & \cellcolor{Ocean}{94}    & \cellcolor{Ocean}{\textbf{95}} \\
    \midrule
    \midrule
    \multicolumn{1}{c}{\multirow{18}[4]{*}{\textbf{News}}} & Style Transformer & 42    & 83    & 65  \\
          & RALoCN & 47    & 82    & 67  \\
          & StyleLM & 50    & 81    & 68  \\
          & SDR   & 55    & 85    & 70  \\
          & DRAG  & 60    & 87    & 72  \\
          & StoryTrans & 68    & 90    & 74  \\
\cmidrule{2-5}          & Role-play+ChatGLM & 72    & 84    & 77  \\
          & Read+ChatGLM & 75    & 85    & 78  \\
          & \cellcolor{Ocean}{\textbf{CAT+ChatGLM}} & \cellcolor{Ocean}{80}    & \cellcolor{Ocean}{86}    & \cellcolor{Ocean}{80}  \\
          & Role-play+Baichuan & 73    & 82    & 75  \\
          & Read+Baichuan & 70    & 80    & 74  \\
          & \cellcolor{Ocean}{\textbf{CAT+Baichuan}} & \cellcolor{Ocean}{83}    & \cellcolor{Ocean}{85}    & \cellcolor{Ocean}{76}  \\
          & Role-play+GPT3.5 & 87    & 89    & 85  \\
          & Read+GPT3.5 & 90    & 91    & 85  \\
          & \cellcolor{Ocean}{\textbf{CAT+GPT3.5}} & \cellcolor{Ocean}{\textbf{95}} & \cellcolor{Ocean}{\textbf{97}} & \cellcolor{Ocean}{88}  \\
          & Role-play+GPT4 & 92    & 95    & 92  \\
          & Read+GPT4 & 94    & 96    & 93  \\
          & \cellcolor{Ocean}{\textbf{CAT+GPT4}} & \cellcolor{Ocean}{93}    & \cellcolor{Ocean}{95}    & \cellcolor{Ocean}{\textbf{95}} \\
    \midrule
    \midrule
    \multicolumn{1}{c}{\multirow{18}[4]{*}{\textbf{Wikipedia}}} & Style Transformer & 40    & 80    & 65  \\
          & RALoCN & 45    & 78    & 67  \\
          & StyleLM & 48    & 77    & 68  \\
          & SDR   & 52    & 81    & 70  \\
          & DRAG  & 57    & 83    & 72  \\
          & StoryTrans & 65    & 86    & 74  \\
\cmidrule{2-5}          & Role-play+ChatGLM & 70    & 80    & 77  \\
          & Read+ChatGLM & 75    & 81    & 78  \\
          & \cellcolor{Ocean}{\textbf{CAT+ChatGLM}} & \cellcolor{Ocean}{80}    & \cellcolor{Ocean}{82}    & \cellcolor{Ocean}{80}  \\
          & Role-play+Baichuan & 73    & 79    & 75  \\
          & Read+Baichuan & 70    & 77    & 74  \\
          & \cellcolor{Ocean}{\textbf{CAT+Baichuan}} & \cellcolor{Ocean}{83}    & \cellcolor{Ocean}{81}    & \cellcolor{Ocean}{76}  \\
          & Role-play+GPT3.5 & 87    & 86    & 85  \\
          & Read+GPT3.5 & 90    & 88    & 85  \\
          & \cellcolor{Ocean}{\textbf{CAT+GPT3.5}} & \cellcolor{Ocean}{\textbf{96}} & \cellcolor{Ocean}{92}    & \cellcolor{Ocean}{88}  \\
          & Role-play+GPT4 & 92    & 93    & 92  \\
          & Read+GPT4 & 94    & \textbf{94} & 93  \\
          & \cellcolor{Ocean}{\textbf{CAT+GPT4}} & \cellcolor{Ocean}{94}    & \cellcolor{Ocean}{93}    & \cellcolor{Ocean}{\textbf{95}} \\
    \bottomrule
    \end{tabular}}
  \label{tab4}%
\end{table}

\subsection{Human Evaluation}

Considering the inherent limitations of automatic evaluation methods, such as BERTScore and BLEU, which often exhibit mechanical assessment tendencies and fixed evaluation standards, these methods cannot fully capture the nuances of language style. To address this, we conducted a comprehensive human assessment of our proposed models and other baseline models. In collaboration with three PhD students in literature, we assessed texts generated by 18 transfer methods across the three datasets, using criteria from \cite{yi2021}. The three datasets are consistent with those presented in the automatic evaluation, facilitating a comprehensive assessment of the models, with specific details shown in Table \ref{tab4}. Our evaluation focused on transfer accuracy, content preservation, and fluency, with potential scores from 0 (the worst) to 100 (the best). The human evaluation results of the transferred texts for all models on other parts of the dataset can be referred in Appendix Tables \ref{tabA6}-\ref{tabA8}.

In particular, CAT-based models (CAT+ChatGLM, CAT+Baichuan, CAT+GPT-3.5-Turbo, and CAT+GPT-4) exhibit superior performance across all metrics compared to their counterparts. For instance, CAT+GPT-3.5-Turbo achieves the highest transfer accuracy scores of 95 and 96 on ``News'' and ``Wikipedia'' datasets, respectively. This can be attributed to GPT-3.5-Turbo's advanced capability to understand and internalize the style definitions summarized by the TSD module, leading to a more accurate execution of the style transfer task. These models also excel in content preservation, with CAT+GPT-3.5-Turbo scoring about 90, indicating their effectiveness in maintaining the original content while accurately transferring the target style. Furthermore, these models achieve high fluency scores, with CAT+GPT-4 reaching up to 95, demonstrating their ability to generate coherent and fluent language outputs. This consistent outperformance highlights the robustness of the CAT architecture in handling complex style transfer tasks.

% Table generated by Excel2LaTeX from sheet 'abolish'
\begin{table}[b]
  \centering
  \renewcommand\arraystretch{1.2}
  \caption{Ablation study results on the ``The Scream'', and ``Scientific Literature'' datasets.}
  \resizebox{\textwidth}{!}{
    \begin{tabular}{ccccccccccccc}
    \toprule
    \multirow{2}[4]{*}{\textbf{Chinese Text}} & \multirow{2}[4]{*}{\textbf{Style Arrangement}} & \multicolumn{5}{c}{\textbf{Transfer Accuracy(\%)}} &       & \multicolumn{5}{c}{\textbf{Content Preservation(\%)}} \\
\cmidrule{3-7}\cmidrule{9-13}          &       & \textbf{BLEU-1} & \textbf{BLEU-2} & \textbf{Precision} & \textbf{Recall} & \textbf{F1} &       & \textbf{BLEU-1} & \textbf{BLEU-2} & \textbf{Precision} & \textbf{Recall} & \textbf{F1} \\
    \midrule
    \midrule
    \multirow{4}[2]{*}{\textbf{The Scream}} & Sentence & 15.33  & 10.41  & 53.14  & 52.80  & 52.97  &       & \textbf{91.94} & \textbf{86.42} & 95.30  & 95.79  & 95.54  \\
          & Word  & 18.78  & 12.95  & 56.78  & 58.16  & 57.46  &       & 89.62  & 83.03  & 92.95  & 92.68  & 92.81  \\
          & Sentence+Word & 25.55  & 16.97  & 62.11  & 62.96  & 62.53  &       & 86.38  & 78.29  & 91.66  & 90.22  & 90.93  \\
          & \cellcolor{Ocean}{\textbf{Word+Sentence}} & \cellcolor{Ocean}{\textbf{46.78}}  & \cellcolor{Ocean}{\textbf{26.53}}  & \cellcolor{Ocean}{\textbf{79.41}}  & \cellcolor{Ocean}{\textbf{80.36}}  & \cellcolor{Ocean}{\textbf{79.92}}  &  \cellcolor{Ocean}{}     & \cellcolor{Ocean}{88.89}  & \cellcolor{Ocean}{84.59}  & \cellcolor{Ocean}{\textbf{96.16}} & \cellcolor{Ocean}{\textbf{96.42}} & \cellcolor{Ocean}{\textbf{96.31}} \\
    \midrule
    \midrule
    \multirow{4}[2]{*}{\textbf{Scientific Literature}} & Sentence & 17.82  & 12.88  & 56.91  & 57.25  & 57.08  &       & 85.72  & \textbf{81.39} & 92.08  & 91.64  & 91.86  \\
          & Word  & 22.12  & 15.97  & 60.38  & 61.59  & 60.98  &       & \textbf{87.14} & 80.76  & 90.43  & 90.12  & 90.27  \\
          & Sentence+Word & 30.27  & 24.84  & 70.03  & 70.46  & 70.24  &       & 81.52  & 76.14  & 89.31  & 87.79  & 88.55  \\
          & \cellcolor{Ocean}{\textbf{Word+Sentence}} & \cellcolor{Ocean}{\textbf{51.38}} & \cellcolor{Ocean}{\textbf{35.55}} & \cellcolor{Ocean}{\textbf{84.31}} & \cellcolor{Ocean}{\textbf{85.13}} & \cellcolor{Ocean}{\textbf{84.70}} &   \cellcolor{Ocean}{}    & \cellcolor{Ocean}{80.82}  & \cellcolor{Ocean}{73.51}  & \cellcolor{Ocean}{\textbf{95.17}} & \cellcolor{Ocean}{\textbf{94.64}} & \cellcolor{Ocean}{\textbf{94.89}} \\
    \bottomrule
    \end{tabular}}
  \label{tab5}%
\end{table}%

\subsection{Ablation Study}

To further investigate the role and interaction of word and sentence level style definitions in the TSD module, we conducted ablation experiments on the ``The Scream'' and ``Scientific Literature'' datasets using GPT-3.5-Turbo. As shown in Table \ref{tab5}, word level prompts have a more significant positive impact on LLMs' understanding of text style than sentence level prompts. This may be attributed to the fact that word level prompts provide more granular and localized style information, enabling LLMs to more precisely capture subtle semantic nuances and stylistic variations in the text. In contrast, sentence level prompts may be more macroscopic and struggle to convey the finer stylistic differences within the text.

The ablation study further reveals that combining both word level and sentence level prompts yields the best performance, with the ``Words+Sentences'' arrangement achieving the highest scores in both transfer accuracy and content preservation. Specifically, the ``Words+Sentences'' combination results in a transfer accuracy BERT-F1 score of about 80\% and a content preservation BERT-F1 score of about 95\%, outperforming other configurations. This suggests that the synergistic effect of initially focusing on fine-grained semantics through word level prompts, followed by sentence level prompts, enhances the model's ability to maintain both the style and content of the original text. Moreover, placing word level prompts before sentence level prompts yields better results in style transfer experiments compared to the reverse order. This may be because word level prompts can guide the LLMs to initially focus on fine-grained semantics and local stylistic information in the text. This arrangement can learn and retain the subtle semantic differences in the text more effectively, thereby more accurately conveying the required style and enhancing the sensitivity of the LLMs to the overall style of the text.

These findings underscore how crucial prompt arrangement is in style transfer tasks. By combining both granular word level and holistic sentence level style cues, we can achieve more accurate and effective results in text generation models. This approach leverages the detailed, fine-grained semantics provided by word level prompts and the broader context captured by sentence level prompts, resulting in a more nuanced and precise style transfer. Such a strategy not only enhances the model's ability to preserve content but also ensures that the stylistic nuances of the original text are maintained, leading to superior overall performance.

\subsection{Case Study}

\begin{figure}[t]
  \centering
  \includegraphics[width=\textwidth]{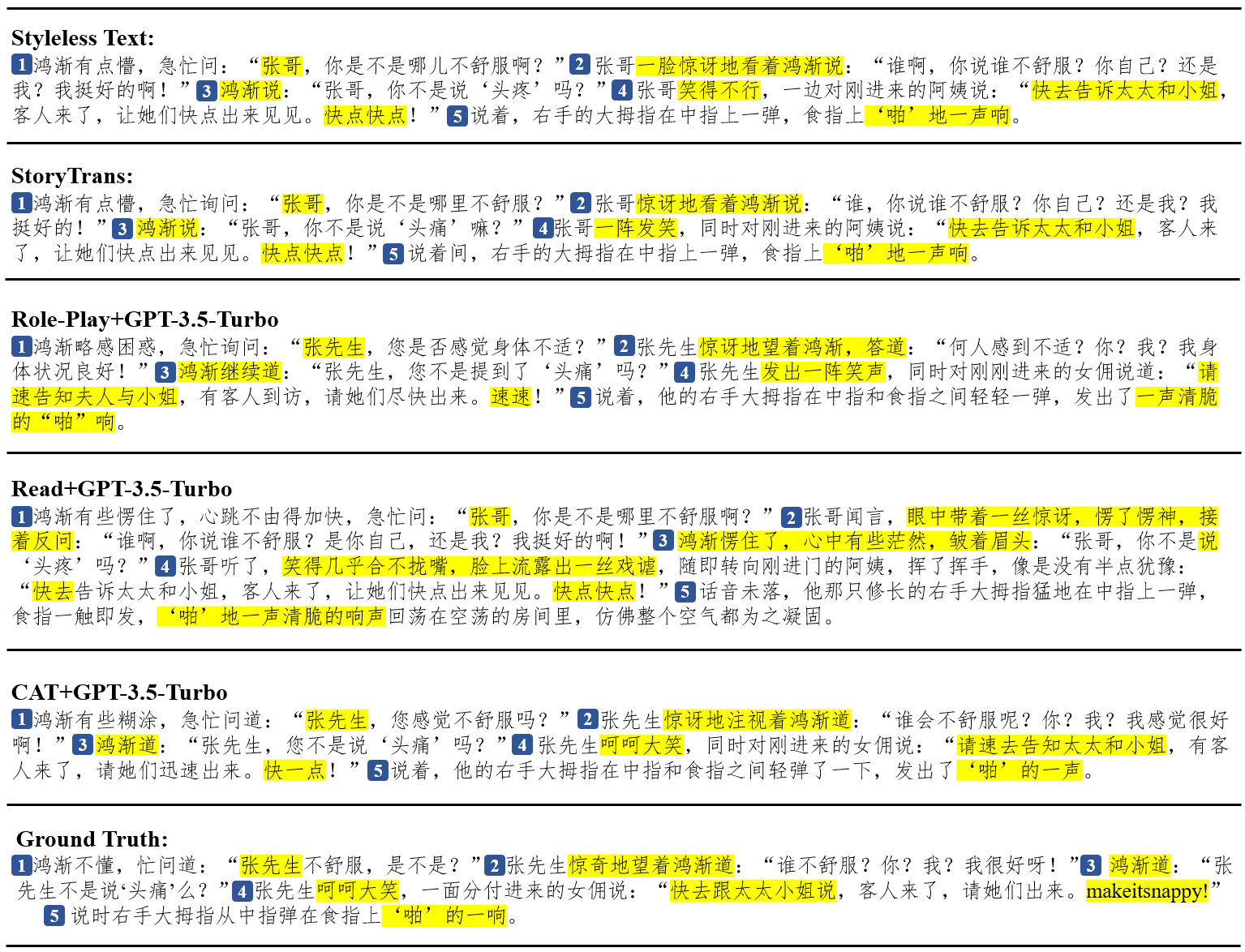}
   \caption{Case study on the ``Fortress Besieged'' dataset.}
   \label{fig5}
\end{figure}

In this case study, Figure \ref{fig5} shows different types of text generated in the experiment of the ``Fortress Besieged'' dataset, including Styleless Text, Storytrans,  Role-Play+GPT-3.5-Turbo, Read+GPT-3.5-Turbo, and CAT+GPT-3.5-Turbo, and Ground Truth. For the demonstration of case studies on other datasets, please refer to Figures \ref{figA12}-\ref{figA17} in the Appendix. We marked the words and sentences that can represent the text style with a yellow background to facilitate our observation of the models‘ transfer effect. Our analysis focuses particularly on the performance of CAT+GPT-3.5-Turbo, which demonstrates exceptional accuracy in style transfer and preservation of original content.

CAT+GPT-3.5-Turbo not only retains the relevant information from the Styleless Text but also closely matches the style of the Ground Truth. The Ground Truth passage features a dialogue between two characters, reflecting a humorous and light-hearted atmosphere. The text generated by CAT+GPT-3.5-Turbo accurately captures this atmosphere and maintains a very natural linguistic style. For instance, ``Mr. Zhang looked at Hongjian with surprise and said: ‘Who would be uncomfortable? You? Me? I feel very good!’'' and ``Mr. Zhang laughed and said to the maid who had just entered: ‘Please go tell the wife and Miss that there are guests coming, and ask them to come out quickly. Hurry up!’'' demonstrate the natural flow of conversation and the accurate portrayal of the characters' personalities. In contrast, texts generated by other models are somewhat lacking in maintaining the original intent and style. CAT+GPT-3.5-Turbo excels not only in precise word choice but also in better reflecting the tone and style of the original text, demonstrating a high level of text style transfer capability and completeness of original content. In general, CAT+GPT-3.5-Turbo showcases significant advantages in text style transfer in this case study, making it an excellent tool for research and practical applications.

\subsection{Stylistic Feature Visualization}

\begin{figure}[t]
\centering
\subfigure[Styleless text and Ground truth text.]{
\label{fig6.1}
\includegraphics[width=0.49\textwidth]{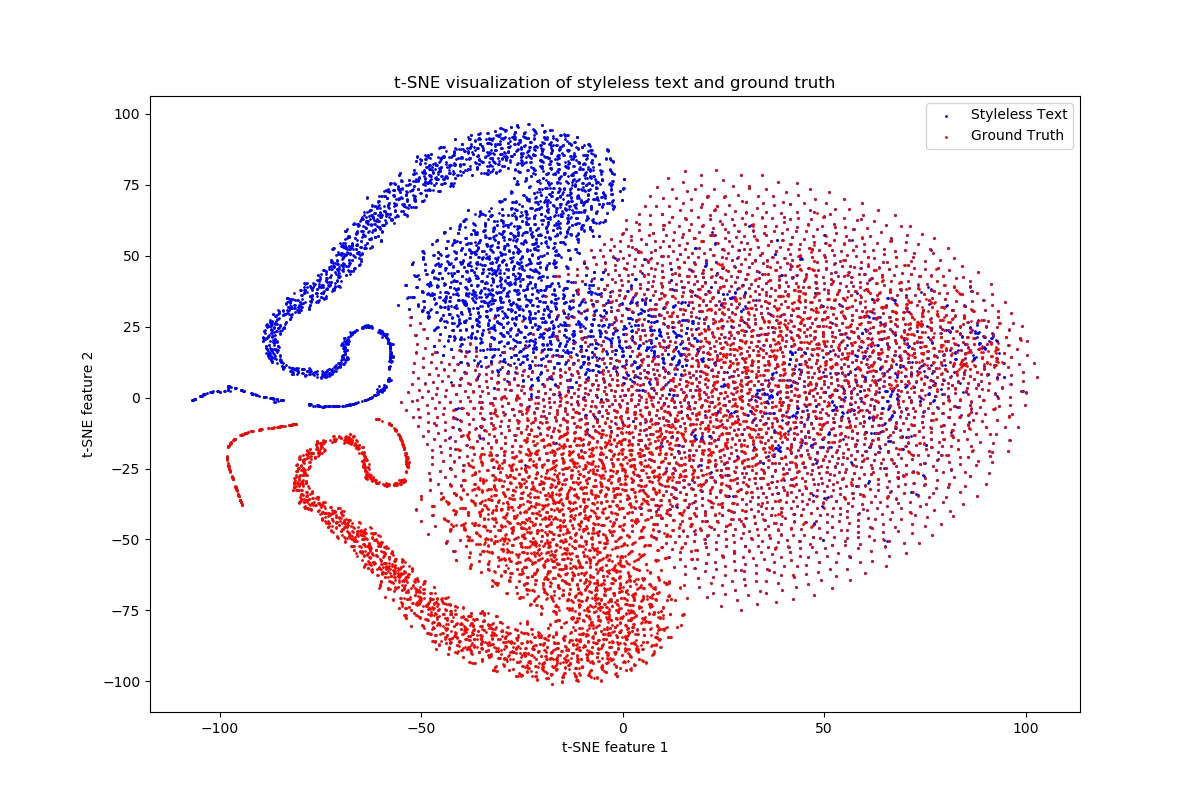}}
\subfigure[Styleless text and Role-play+GPT-3.5-Turbo text.]{
\label{fig6.2}
\includegraphics[width=0.49\textwidth]{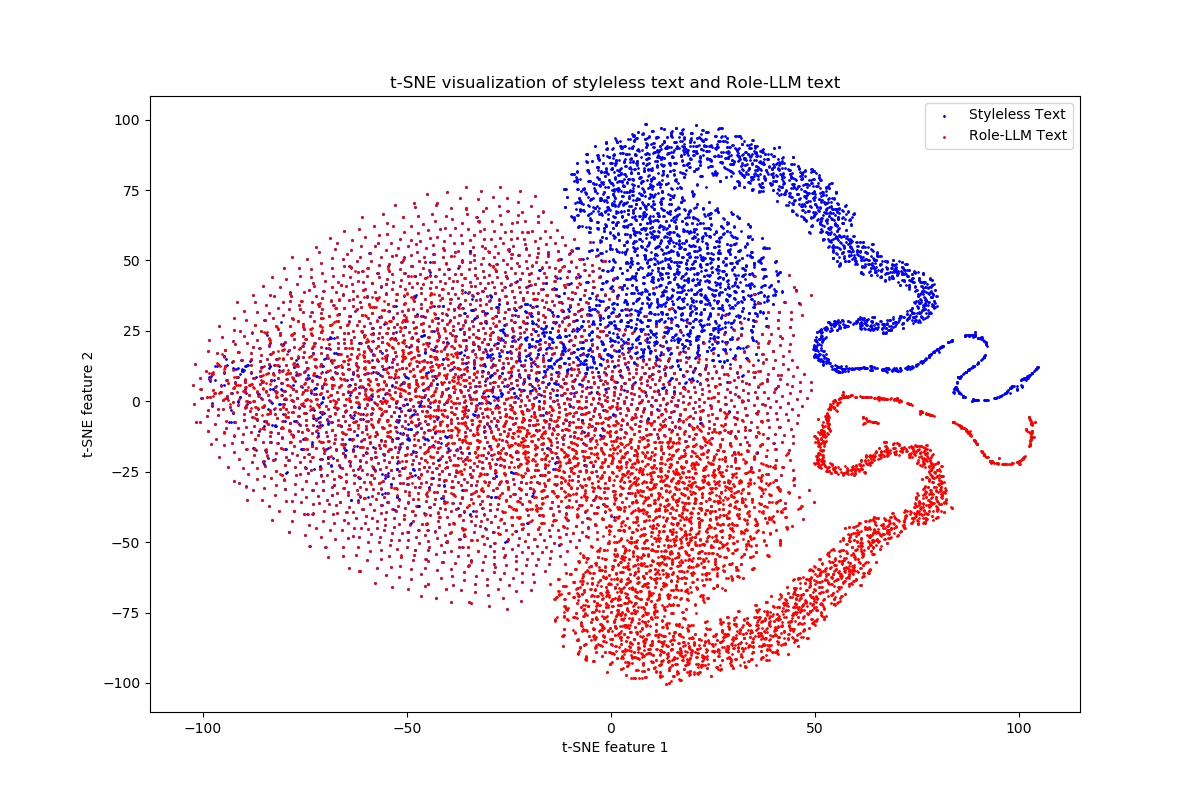}}

\subfigure[Styleless text and Read+GPT-3.5-Turbo text.]{
\label{fig6.3}
\includegraphics[width=0.49\textwidth]{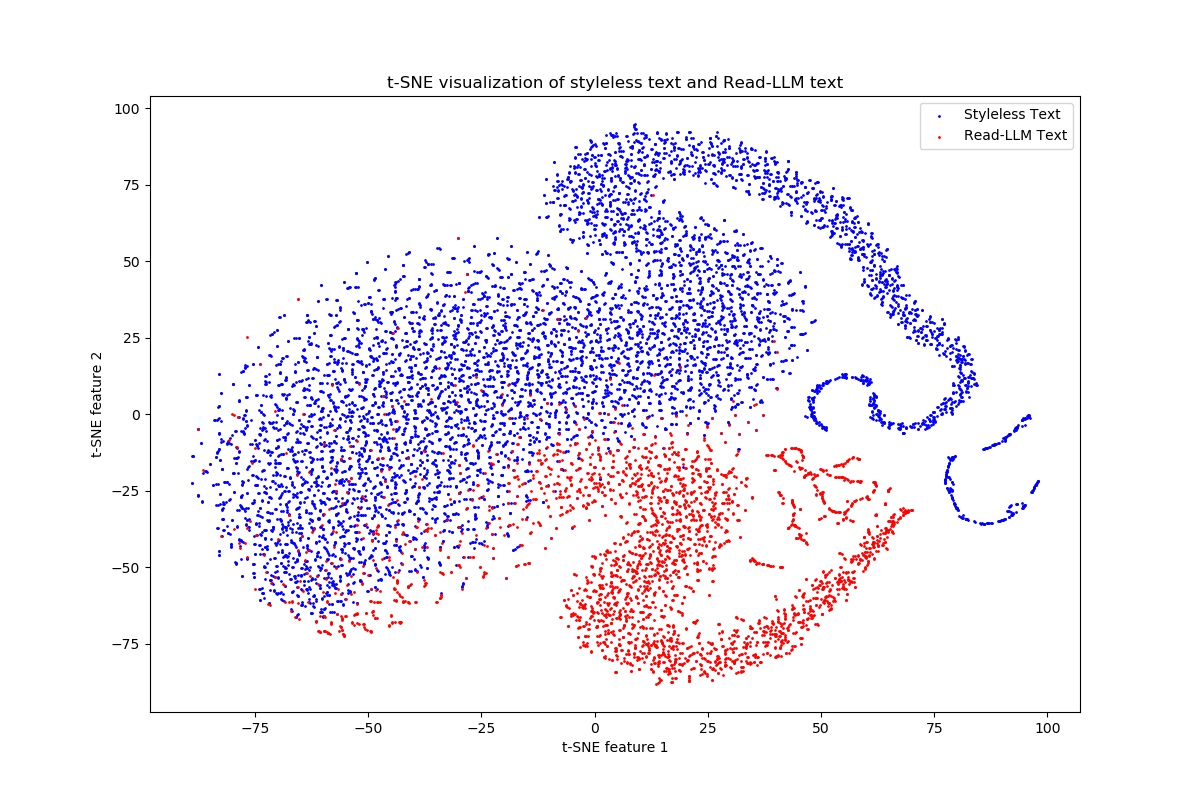}}
\subfigure[Styleless text and CAT+GPT-3.5-Turbo text.]{
\label{fig6.4}
\includegraphics[width=0.49\textwidth]{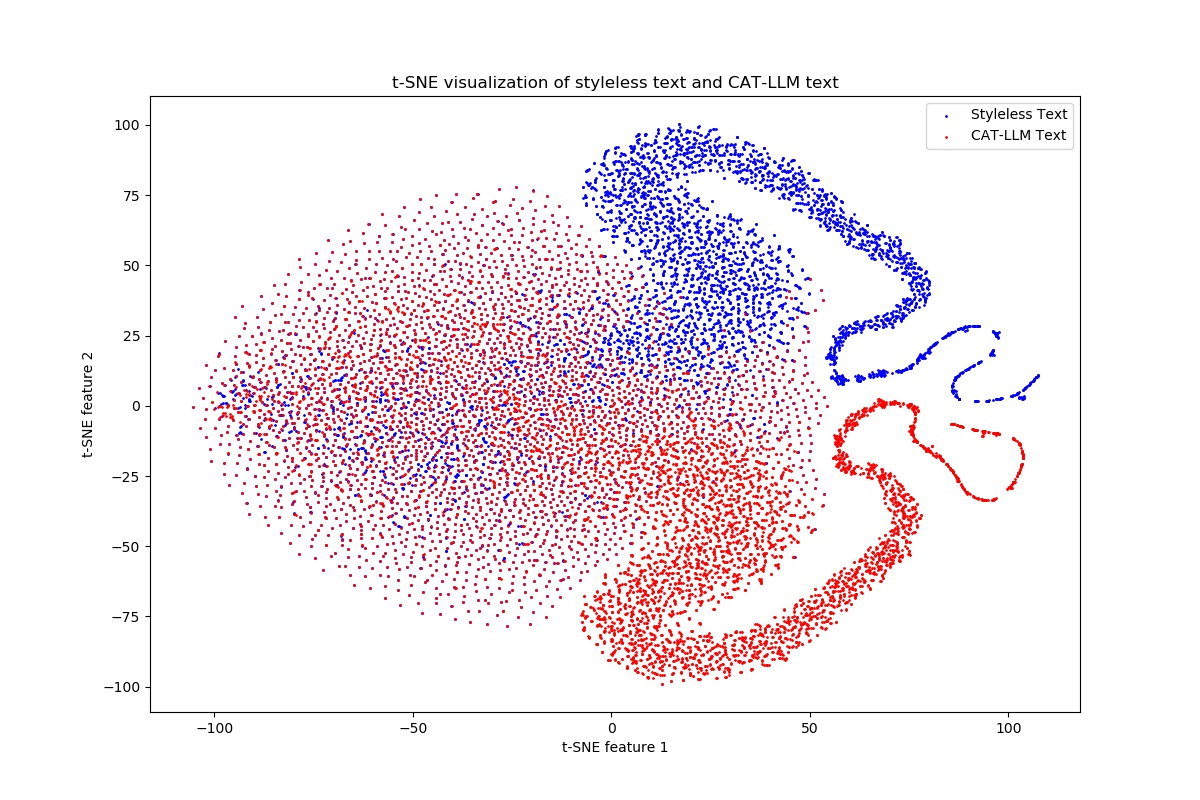}}
\caption{Stylistic features visualization of styleless text and transferred text on the ``Fortress Besieged'' dataset using t-SNE.}
\label{fig6}
\end{figure}

In our research on stylistic transfer, we follow the approach outlined by \cite{zhu-etal} to define and visualize stylistic features. Utilizing t-SNE for visualization, we provide a detailed comparative analysis of the stylistic features present in different sets of styleless texts and transferred texts, as shown in Figure \ref{fig6}. Specifically, we created four figures to illustrate these comparisons: (1) Styleless text and Ground truth text, (2) Styleless text and Role-play + GPT-3.5-Turbo text, (3) Styleless text and Read + GPT-3.5-Turbo text, and (4) Styleless text and CAT + GPT-3.5-Turbo text. The fourth figure, which compares the styleless text with the CAT + GPT-3.5-Turbo generated text, shows the closest resemblance to the Ground truth text. This suggests that the CAT + GPT-3.5-Turbo model does an excellent job of capturing and transferring the specific stylistic features of the source texts. In Figure \ref{fig6.1} and \ref{fig6.4}, you can see that the stylistic elements in texts generated by CAT + GPT-3.5-Turbo closely match those in the Ground truth texts. Texts generated by the Role-play + GPT-3.5-Turbo model also perform well, showing a slight difference compared to the CAT + GPT-3.5-Turbo model but still maintaining a high level of similarity to the Ground truth text. In contrast, the Read + GPT-3.5-Turbo model shows a more noticeable deviation in these stylistic features compared to the Ground truth texts. This indicates that while Role-play + GPT-3.5-Turbo can adapt styles almost as effectively as CAT + GPT-3.5-Turbo, the Read + GPT-3.5-Turbo model lags significantly behind. These findings are consistent with our automatic evaluation experiments, which also indicate that the CAT + GPT-3.5-Turbo model outperforms the other models in stylistic transfer.

\subsection{Discussion}

This study emphasizes the significant advantages of the CAT-LLM framework in achieving precise text style transfer, which comprehensively integrates word and sentence level style extraction algorithms. This clever combination enables the model to strike a delicate balance between preserving the original content and transforming its stylistic expression. By adopting this fine-grained approach, CAT-LLM is able to capture the inherent subtle differences in various Chinese text styles, including classical Chinese novels and modern social media texts. Through the meticulous design of prompts, CAT-LLM not only enhances stylistic consistency but also outperforms traditional methods and recent large language models in maintaining content integrity. The results of comprehensive evaluations, including automated assessments and human scoring, demonstrate the framework's ability to generate text that is stylistically accurate and content-preserving. These findings suggest that CAT-LLM holds considerable promise as a valuable tool for more sophisticated text generation tasks.

In addition to its strong performance in text style transfer, CAT-LLM is designed with real-world deployment and cost-efficiency in mind. Its modular architecture and the core Text Style Definition (TSD) module allow for flexible integration into various applications, such as digital media, content creation, and customer communication. The model’s ability to deliver high-quality, stylistically consistent output can be leveraged across industries, from enhancing the productivity of journalists and writers to adapting educational content for modern audiences. Moreover, CAT-LLM is computationally efficient, relying on smaller, pretrained models like ChatGLM (6B) for fine-tuning, reducing the computational costs typically associated with large language models. This approach, combined with cost-effective third-party APIs like GPT-3.5-Turbo and GPT-4, ensures that CAT-LLM remains accessible to a wide range of industries. Looking ahead, future improvements such as knowledge distillation and lightweight fine-tuning will further enhance its scalability and affordability, making CAT-LLM a practical and sustainable solution for diverse text generation tasks.

While the CAT-LLM framework shows promising results in text style transfer, there are several limitations that warrant further exploration. One key limitation is that the current model is optimized for Chinese text, which restricts its applicability to other languages. Expanding its capabilities to support multilingual text styles would significantly broaden its potential. Additionally, although the model effectively preserves content integrity, challenges remain in minimizing the introduction of irrelevant or hallucinated content, especially when dealing with more complex stylistic nuances. Future research should also aim to refine evaluation metrics by incorporating factors such as readability and creativity, which would offer a deeper understanding of the generated text's quality. Finally, integrating domain-specific knowledge into style prompts could improve precision in specialized fields such as technical writing or literary analysis, improving the applicability of CAT-LLM to various tasks.

\section{Conclusion and Future Work}

In this paper, we propose the \textbf{C}hinese \textbf{A}rticle-style \textbf{T}ransfer (\textbf{CAT-LLM}) framework , a novel approach to addressing the challenges of style transfer in complex Chinese long texts. By integrating a specialized Text Style Definition (TSD) module, CAT-LLM effectively analyzes and adapts to the nuanced stylistic features of Chinese texts at both the word and sentence levels. This module bridges the gap between large language models (LLMs) and the intricate rhetorical, structural elements of Chinese writing. To facilitate robust evaluation, we constructed ten parallel datasets using ChatGPT and various Chinese texts, each corresponding to distinct styles of Chinese articles. These datasets provide a novel evaluation paradigm for style transfer research. Extensive experimental results show that CAT-LLM achieves a transfer accuracy F1 score of approximately 80\% and a content preservation F1 score of around 95\% across most datasets, significantly outperforming existing methods in balancing style transfer accuracy with content preservation.

Despite the promising results demonstrated by CAT-LLM, there are still several areas that require further refinement. Future work will focus on expanding the framework to support multilingual text style transfer, which would allow the model to process a wider range of languages and enhance its applicability. Additionally, addressing challenges related to content preservation while adapting more complex stylistic features, such as emotional tones and domain-specific attributes, will be a priority. To improve the evaluation process, future research will incorporate additional metrics, such as readability and creativity, to provide a more comprehensive assessment of the model's performance. Lastly, integrating domain-specific knowledge into style prompts will help to enhance the performance of CAT-LLM in specialized applications, such as technical writing or literary analysis, ensuring its effectiveness in a wider range of tasks.

\begin{acks}
This work was supported by the National Natural Science Foundation of China (72271233), the Fundamental Research Funds for the Central Universities (2024110591), Beijing Municipal Science and Technology Project (Z241100001324020), Suzhou Key Laboratory of Artificial Intelligence and Social Governance Technologies (SZS2023007), Smart Social Governance Technology and Innovative Application Platform (YZCXPT2023101), and The Innovation System of the Integration between Industry and Education for Smart Governance (CJRH2024101).
\end{acks}

\bibliographystyle{ACM-Reference-Format}
\bibliography{sample-base}

\newpage
\appendix

\renewcommand{\thefigure}{A\arabic{figure}}
\setcounter{figure}{0}
\renewcommand{\thetable}{A\arabic{table}}
\setcounter{table}{0}

\section{Appendix}

\subsection{Experiment Results, Style Definitions, and Case studies}

% Table generated by Excel2LaTeX from sheet 'The Scream'
\begin{table}[htbp]
  \centering
  \caption{Automatic evaluation results on ``The Scream''.}
  \resizebox{\textwidth}{!}{
    \begin{tabular}{ccccccccccccc}
    \toprule
    \multirow{2}[4]{*}{\textbf{Chinese Text}} & \multirow{2}[4]{*}{\textbf{Models}} & \multicolumn{5}{c}{\textbf{Transfer Accuracy(\%)}} &       & \multicolumn{5}{c}{\textbf{Content Preservation(\%)}} \\
\cmidrule{3-7}\cmidrule{9-13}          &       & \textbf{BLEU-1} & \textbf{BLEU-2} & \textbf{Precision} & \textbf{Recall} & \textbf{F1} &       & \textbf{BLEU-1} & \textbf{BLEU-2} & \textbf{Precision} & \textbf{Recall} & \textbf{F1} \\
    \midrule
    \midrule
    \multicolumn{1}{c}{\multirow{18}[4]{*}{\textbf{The Scream}}} & Style Transformer & 16.89  & 10.22  & 59.58  & 60.71  & 60.13  &       & \textbf{92.71} & \textbf{90.38} & \textbf{98.23} & \textbf{98.13} & \textbf{98.18} \\
          & RALoCN & 24.72  & 17.14  & 66.09  & 65.48  & 65.78  &       & 89.02  & 87.16  & 95.12  & 95.85  & 95.49  \\
          & StyleLM & 19.34  & 16.18  & 55.61  & 56.05  & 55.86  &       & 92.31  & 87.81  & 93.92  & 95.22  & 94.82  \\
          & SDR   & 30.23  & 22.15  & 59.67  & 62.66  & 60.48  &       & 92.16  & 85.17  & 92.81  & 91.91  & 92.19  \\
          & DRAG  & 33.14  & 22.52  & 65.18  & 63.29  & 64.33  &       & 90.68  & 86.40  & 95.96  & 95.02  & 95.44  \\
          & StoryTrans & \textbf{33.33}  & \textbf{24.42}  & \textbf{67.81}  & \textbf{69.09}  & \textbf{68.45}  &       & 89.36  & 84.24  & 94.59  & 93.53  & 93.96  \\
\cmidrule{2-7}\cmidrule{9-13}          & Role-play+ChatGLM & 38.67  & 19.87  & 76.89  & 77.12  & 76.99  &       & 75.62  & 68.16  & 88.64  & 89.56  & 89.06  \\
          & Read+ChatGLM & 13.72  & 4.17  & 57.88  & 62.96  & 60.30  &       & 13.89  & 4.82  & 57.99  & 62.84  & 60.30  \\
          & \textbf{CAT+ChatGLM} & 40.44  & 22.47  & 76.17  & 77.87  & 76.98  &       & 74.43  & 67.70  & 91.12  & 92.26  & 91.63  \\
          & Role-play+Baichuan & 36.07  & 16.93  & 75.24  & 76.22  & 75.72  &       & 80.90  & 78.51  & 88.37  & 88.17  & 88.27  \\
          & Read+Baichuan & 17.14  & 7.05  & 59.32  & 64.18  & 61.64  &       & 21.54  & 12.78  & 61.24  & 65.85  & 63.44  \\
          & \textbf{CAT+Baichuan} & 42.55  & 24.31  & 77.32  & 77.65  & 77.45  &       & 83.33  & 79.16  & 94.03  & 93.13  & 93.54  \\
          & Role-play+GPT3.5 & 41.13  & 22.91  & 77.60  & 78.45  & 78.01  &       & 73.82  & 66.03  & 92.42  & 92.27  & 92.34  \\
          & Read+GPT3.5 & 19.76  & 8.94  & 60.91  & 64.75  & 62.74  &       & 29.70  & 22.23  & 64.94  & 68.59  & 66.68  \\
          & \textbf{CAT+GPT3.5} & \textbf{46.78} & \textbf{26.53} & \textbf{79.41} & \textbf{80.36} & \textbf{79.92} &       & \textbf{88.89} & \textbf{84.59} & \textbf{96.16} & \textbf{96.42} & \textbf{96.31} \\
          & Role-play+GPT4 & 34.89  & 16.81  & 72.04  & 74.13  & 73.05  &       & 42.05  & 25.52  & 78.03  & 79.68  & 78.82  \\
          & Read+GPT4 & 30.33  & 14.14  & 73.95  & 73.64  & 73.78  &       & 37.98  & 23.25  & 79.90  & 78.62  & 79.23  \\
          & \textbf{CAT+GPT4} & 36.92  & 18.16  & 73.83  & 75.25  & 74.31  &       & 55.76  & 39.96  & 87.46  & 87.38  & 87.42  \\
    \bottomrule
    \end{tabular}}
  \label{tabA1}%
\end{table}%

% Table generated by Excel2LaTeX from sheet 'Fifteen Year of the Wanli Era'
\begin{table}[htbp]
  \centering
  \caption{Automatic evaluation results on ``Fifteen Year of the Wanli Era''.}
  \resizebox{\textwidth}{!}{
    \begin{tabular}{ccccccccccccc}
    \toprule
    \multirow{2}[4]{*}{\textbf{Chinese Text}} & \multirow{2}[4]{*}{\textbf{Models}} & \multicolumn{5}{c}{\textbf{Transfer Accuracy(\%)}} &       & \multicolumn{5}{c}{\textbf{Content Preservation(\%)}} \\
\cmidrule{3-7}\cmidrule{9-13}          &       & \textbf{BLEU-1} & \textbf{BLEU-2} & \textbf{Precision} & \textbf{Recall} & \textbf{F1} &       & \textbf{BLEU-1} & \textbf{BLEU-2} & \textbf{Precision} & \textbf{Recall} & \textbf{F1} \\
    \midrule
    \midrule
    \multirow{18}[4]{*}{\textbf{Fifteen Year of the Wanli Era}} & Style Transformer & 14.60  & 10.01  & 52.10  & 51.76  & 51.93  &       & \textbf{93.82} & \textbf{88.18} & \textbf{94.36} & \textbf{94.84} & \textbf{94.60} \\
          & RALoCN & 18.62  & 13.58  & 58.01  & 58.41  & 58.31  &       & 92.89  & 86.79  & 89.49  & 88.92  & 89.20  \\
          & StyleLM & 17.89  & 12.45  & 55.67  & 57.02  & 56.33  &       & 91.45  & 84.72  & 92.03  & 91.76  & 91.92  \\
          & SDR   & 24.33  & 16.32  & 60.89  & 61.73  & 61.15  &       & 88.14  & 79.89  & 90.75  & 89.33  & 89.95  \\
          & DRAG  & 26.72  & 18.45  & 63.14  & 64.06  & 63.72  &       & 89.66  & 81.32  & 92.49  & 91.78  & 92.13  \\
          & StoryTrans & \textbf{28.51} & \textbf{20.15} & \textbf{65.22} & \textbf{66.49} & \textbf{65.85} &       & 90.52  & 85.74  & 93.11  & 92.45  & 92.88  \\
\cmidrule{2-7}\cmidrule{9-13}          & Role-play+ChatGLM & 24.37  & 14.39  & 67.90  & 68.13  & 67.99  &       & 74.89  & 65.93  & 91.83  & 90.94  & 91.36  \\
          & Read+ChatGLM & 13.09  & 3.33  & 54.40  & 57.37  & 55.83  &       & 16.67  & 6.04  & 57.69  & 60.76  & 59.17  \\
          & \textbf{CAT+ChatGLM} & 43,62 & 25.37  & 78.63  & 79.60  & 79.08  &       & 85.54  & 79.85  & 93.98  & 93.97  & 93.97  \\
          & Role-play+Baichuan & 37.33  & 21.88  & 74.99  & 79.18  & 76.84  &       & 70.46  & 68.75  & 83.85  & 88.65  & 85.79  \\
          & Read+Baichuan & 18.04  & 6.20  & 59.25  & 62.25  & 60.69  &       & 26.51  & 16.60  & 62.50  & 65.39  & 63.89  \\
          & \textbf{CAT+Baichuan} & 41.88  & 23.99  & 77.62  & 77.77  & 77.66  &       & 90.54  & 89.14  & 95.52  & 95.24  & 95.37  \\
          & Role-play+GPT3.5 & 44.71  & 26.24  & 79.35  & 80.24  & 79.78  &       & 84.38  & 78.70  & 95.67  & 95.51  & 95.58  \\
          & Read+GPT3.5 & 16.13  & 5.76  & 58.55  & 62.19  & 60.30  &       & 23.46  & 14.40  & 61.54  & 65.14  & 63.27  \\
          & \textbf{CAT+GPT3.5} & \textbf{46.27} & \textbf{27.94} & \textbf{80.78} & \textbf{80.66} & \textbf{80.73} &       & \textbf{93.63} & \textbf{90.14} & \textbf{96.17} & \textbf{96.99} & \textbf{96.56} \\
          & Role-play+GPT4 & 30.50  & 14.31  & 69.09  & 71.87  & 70.41  &       & 40.29  & 23.79  & 74.25  & 77.11  & 75.60  \\
          & Read+GPT4 & 29.57  & 13.83  & 72.98  & 74.76  & 73.78  &       & 35.82  & 20.74  & 77.73  & 79.17  & 78.33  \\
          & \textbf{CAT+GPT4} & 36.99  & 19.28  & 75.73  & 76.53  & 76.12  &       & 56.49  & 49.96  & 84.37  & 84.41  & 84.39  \\
    \bottomrule
    \end{tabular}}
  \label{tabA2}%
\end{table}%

% Table generated by Excel2LaTeX from sheet 'Zeng guo fan'
\begin{table}[htbp]
  \centering
  \caption{Automatic evaluation results on ``The Family Instructions of Zeng Guofan''.}
  \resizebox{\textwidth}{!}{
    \begin{tabular}{ccccccccccccc}
    \toprule
    \multirow{2}[4]{*}{\textbf{Chinese Text}} & \multirow{2}[4]{*}{\textbf{Models}} & \multicolumn{5}{c}{\textbf{Transfer Accuracy(\%)}} &       & \multicolumn{5}{c}{\textbf{Content Preservation(\%)}} \\
\cmidrule{3-7}\cmidrule{9-13}          &       & \textbf{BLEU-1} & \textbf{BLEU-2} & \textbf{Precision} & \textbf{Recall} & \textbf{F1} &       & \textbf{BLEU-1} & \textbf{BLEU-2} & \textbf{Precision} & \textbf{Recall} & \textbf{F1} \\
    \midrule
    \midrule
    \multirow{18}[4]{*}{\textbf{The Family Instructions of Zeng Guofan}} & Style Transformer & 12.98  & 8.42  & 56.76  & 59.42  & 58.07  &       & \textbf{92.30} & \textbf{91.44} & \textbf{98.27} & \textbf{98.33} & \textbf{98.30} \\
          & RALoCN & 21.87  & 15.17  & 58.23  & 58.04  & 58.13  &       & 90.20  & 88.33  & 95.23  & 95.89  & 95.74  \\
          & StyleLM & 19.75  & 12.35  & 57.21  & 57.82  & 57.51  &       & 91.55  & 89.72  & 96.21  & 96.53  & 96.34  \\
          & SDR   & 27.50  & 18.15  & 62.10  & 63.15  & 62.48  &       & 88.31  & 84.55  & 94.45  & 95.20  & 84.86  \\
          & DRAG  & 30.10  & 20.45  & 64.40  & 65.93  & 65.27  &       & 89.65  & 86.11  & 95.75  & 96.32  & 95.97  \\
          & StoryTrans & \textbf{31.45} & \textbf{22.12} & \textbf{66.75} & \textbf{67.80} & \textbf{67.14} &       & 90.45  & 87.35  & 96.78  & 97.05  & 96.93  \\
\cmidrule{2-7}\cmidrule{9-13}          & Role-play+ChatGLM & 40.30  & 21.56  & 75.38  & 74.90  & 75.11  &       & 66.19  & 55.29  & 87.97  & 86.97  & 87.43  \\
          & Read+ChatGLM & 19.05  & 5.20  & 57.05  & 59.02  & 58.01  &       & 19.95  & 6.05  & 57.75  & 59.84  & 58.77  \\
          & \textbf{CAT+ChatGLM} & 41.16  & 23.25  & 76.58  & 77.25  & 76.87  &       & 78.34  & 72.32  & 93.23  & 91.68  & 92.40  \\
          & Role-play+Baichuan & 37.15  & 21.82  & 69.64  & 72.25  & 70.78  &       & 73.66  & 71.97  & 85.06  & 88.96  & 86.60  \\
          & Read+Baichuan & 18.76  & 6.52  & 58.37  & 60.97  & 59.62  &       & 24.25  & 13.41  & 60.73  & 63.59  & 62.10  \\
          & \textbf{CAT+Baichuan} & 41.97  & 23.61  & 76.74  & 76.43  & 76.56  &       & 89.14  & 87.61  & 95.09  & 94.81  & 94.94  \\
          & Role-play+GPT3.5 & 39.35  & 21.57  & 76.88  & 76.53  & 76.68  &       & 69.63  & 60.44  & 91.42  & 90.40  & 90.89  \\
          & Read+GPT3.5 & 14.72  & 4.82  & 56.63  & 60.39  & 58.44  &       & 15.44  & 5.77  & 57.03  & 60.82  & 58.85  \\
          & \textbf{CAT+GPT3.5} & \textbf{45.97} & \textbf{26.94} & \textbf{78.54} & \textbf{79.06} & \textbf{78.79} &       & \textbf{90.72} & \textbf{87.15} & \textbf{97.23} & \textbf{97.41} & \textbf{97.31} \\
          & Role-play+GPT4 & 24.09  & 8.94  & 69.76  & 68.47  & 69.07  &       & 26.50  & 11.49  & 73.19  & 71.61  & 72.36  \\
          & Read+GPT4 & 32.11  & 14.97  & 71.88  & 72.14  & 71.97  &       & 47.09  & 32.57  & 78.56  & 83.83  & 81.03  \\
          & \textbf{CAT+GPT4} & 33.55  & 18.24  & 73.30  & 74.35  & 73.84  &       & 40.84  & 25.14  & 77.33  & 77.40  & 77.32  \\
    \bottomrule
    \end{tabular}}
  \label{tabA3}%
\end{table}%

% Table generated by Excel2LaTeX from sheet 'Santi'
\begin{table}[htbp]
  \centering
  \caption{Automatic evaluation results on ``The Three-Body Problem''.}
  \resizebox{\textwidth}{!}{
    \begin{tabular}{ccccccccccccc}
    \toprule
    \multirow{2}[4]{*}{\textbf{Chinese Text}} & \multirow{2}[4]{*}{\textbf{Models}} & \multicolumn{5}{c}{\textbf{Transfer Accuracy(\%)}} &       & \multicolumn{5}{c}{\textbf{Content Preservation(\%)}} \\
\cmidrule{3-7}\cmidrule{9-13}          &       & \textbf{BLEU-1} & \textbf{BLEU-2} & \textbf{Precision} & \textbf{Recall} & \textbf{F1} &       & \textbf{BLEU-1} & \textbf{BLEU-2} & \textbf{Precision} & \textbf{Recall} & \textbf{F1} \\
    \midrule
    \midrule
    \multirow{18}[4]{*}{\textbf{The Three-Body Problem}} & Style Transformer & 14.45  & 7.39  & 59.11  & 59.27  & 59.19  &       & \textbf{92.00} & \textbf{90.55} & \textbf{97.20} & \textbf{97.91} & \textbf{97.05} \\
          & RALoCN & 16.56  & 9.67  & 59.01  & 59.46  & 59.24  &       & 87.60  & 80.25  & 94.42  & 96.16  & 95.80  \\
          & StyleLM & 18.95  & 11.54  & 58.72  & 58.98  & 58.85  &       & 89.24  & 85.14  & 94.99  & 96.18  & 95.58  \\
          & SDR   & 24.35  & 15.31  & 62.35  & 63.45  & 62.85  &       & 85.55  & 79.47  & 93.49  & 94.17  & 93.86  \\
          & DRAG  & 26.80  & 17.26  & 64.01  & 65.19  & 64.78  &       & 87.16  & 81.74  & 94.18  & 95.35  & 94.81  \\
          & StoryTrans & \textbf{28.69}  & \textbf{19.22}  & \textbf{65.83}  & \textbf{66.85}  & \textbf{66.32}  &       & 88.23  & 83.26  & 95.31  & 93.91  & 94.62  \\
\cmidrule{2-7}\cmidrule{9-13}          & Role-play+ChatGLM & 42.37  & 23.92  & 75.09  & 75.97  & 75.39  &       & 65.48  & 55.58  & 88.30  & 86.90  & 87.53  \\
          & Read+ChatGLM & 15.06  & 5.78  & 58.18  & 62.95  & 60.46  &       & 19.37  & 11.52  & 59.55  & 64.30  & 61.81  \\
          & \textbf{CAT+ChatGLM} & 45.89  & 25.79  & 75.90  & 76.80  & 76.16  &       & 79.71  & 74.13  & 94.07  & 92.48  & 93.21  \\
          & Role-play+Baichuan & 42.29  & 25.52  & 73.87  & 76.43  & 74.99  &       & 77.57  & 75.20  & 89.51  & 92.05  & 90.48  \\
          & Read+Baichuan & 25.59  & 12.80  & 64.39  & 67.90  & 66.06  &       & 40.21  & 32.91  & 70.79  & 73.80  & 72.21  \\
          & \textbf{CAT+Baichuan} & 50.05  & 30.68  & 78.49  & 79.33  & 78.89  &       & 86.52  & 84.73  & 93.44  & 93.07  & 93.24  \\
          & Role-play+GPT3.5 & 45.52  & 27.39  & 77.02  & 77.90  & 77.43  &       & 68.09  & 58.08  & 88.82  & 88.82  & 88.80  \\
          & Read+GPT3.5 & 26.24  & 13.36  & 64.63  & 68.05  & 66.26  &       & 39.53  & 31.66  & 70.50  & 73.58  & 71.95  \\
          & \textbf{CAT+GPT3.5} & \textbf{50.28} & \textbf{30.80} & \textbf{79.24} & \textbf{80.46} & \textbf{79.83} &       & \textbf{88.16} & \textbf{83.80} & \textbf{96.13} & \textbf{96.39} & \textbf{96.25} \\
          & Role-play+GPT4 & 35.34  & 18.01  & 70.07  & 73.98  & 71.95  &       & 40.89  & 24.14  & 74.07  & 77.74  & 75.84  \\
          & Read+GPT4 & 33.05  & 15.88  & 71.20  & 73.04  & 72.08  &       & 37.95  & 21.97  & 75.83  & 77.15  & 76.45  \\
          & \textbf{CAT+GPT4} & 40.11  & 22.05  & 72.80  & 76.72  & 74.69  &       & 51.62  & 36.07  & 79.84  & 83.73  & 81.71  \\
    \bottomrule
    \end{tabular}}
  \label{tabA4}%
\end{table}%

% Table generated by Excel2LaTeX from sheet 'Wenxian'
\begin{table}[htbp]
  \centering
  \caption{Automatic evaluation results on ``Scientific Literature''.}
  \resizebox{\textwidth}{!}{
    \begin{tabular}{ccccccccccccc}
    \toprule
    \multirow{2}[4]{*}{\textbf{Chinese Text}} & \multirow{2}[4]{*}{\textbf{Models}} & \multicolumn{5}{c}{\textbf{Transfer Accuracy(\%)}} &       & \multicolumn{5}{c}{\textbf{Content Preservation(\%)}} \\
\cmidrule{3-7}\cmidrule{9-13}          &       & \textbf{BLEU-1} & \textbf{BLEU-2} & \textbf{Precision} & \textbf{Recall} & \textbf{F1} &       & \textbf{BLEU-1} & \textbf{BLEU-2} & \textbf{Precision} & \textbf{Recall} & \textbf{F1} \\
    \midrule
    \midrule
    \multirow{18}[4]{*}{\textbf{Scientific Literature}} & Style Transformer & 15.33  & 10.41  & 53.14  & 52.80  & 52.97  &       & \textbf{91.94} & \textbf{86.42} & \textbf{95.30} & \textbf{95.79} & \textbf{95.54} \\
          & RALoCN & 19.55  & 14.12  & 59.17  & 59.58  & 59.37  &       & 91.03  & 85.05  & 90.38  & 89.81  & 90.09  \\
          & StyleLM & 18.78 & 12.95 & 56.78 & 58.16 & 57.46 &       & 89.62  & 83.03  & 92.95  & 92.68  & 92.81  \\
          & SDR   & 25.55 & 16.97 & 62.11 & 62.96 & 62.53 &       & 86.38  & 78.29  & 91.66  & 90.22  & 90.93  \\
          & DRAG  & 28.06 & 19.19 & 64.40  & 65.34 & 64.86 &       & 87.87  & 79.69  & 93.41  & 92.70  & 93.05  \\
          & StoryTrans & \textbf{29.94} & \textbf{20.96} & \textbf{66.52} & \textbf{67.82} & \textbf{67.16} &       & 88.71  & 84.03  & 94.04  & 93.37  & 93.70  \\
\cmidrule{2-7}\cmidrule{9-13}          & Role-play+ChatGLM & 47.74  & 32.16  & 81.22  & 82.74  & 81.94  &       & 67.85  & 54.59  & 89.42  & 89.47  & 89.43  \\
          & Read+ChatGLM & 40.34  & 27.06  & 74.86  & 79.66  & 77.12  &       & 59.11  & 51.01  & 82.30  & 87.59  & 84.73  \\
          & \textbf{CAT+ChatGLM} & 48.58  & 32.65  & 81.55  & 82.73  & 82.12  &       & 72.33  & 61.60  & 91.38  & 91.83  & 91.57  \\
          & Role-play+Baichuan & 43.43  & 27.70  & 78.83  & 80.49  & 79.61  &       & 55.91  & 39.87  & 83.37  & 84.04  & 83.66  \\
          & Read+Baichuan & 29.12  & 16.54  & 70.35  & 70.74  & 70.52  &       & 45.64  & 33.45  & 75.72  & 75.70  & 75.69  \\
          & \textbf{CAT+Baichuan} & 45.62  & 30.00  & 80.29  & 81.13  & 80.68  &       & 72.98  & 63.18  & 91.20  & 90.75  & 90.94  \\
          & Role-play+GPT3.5 & 49.64  & 34.11  & 83.08  & 83.19  & 83.12  &       & 67.13  & 54.63  & 90.55  & 89.28  & 89.89  \\
          & Read+GPT3.5 & 48.49  & 32.89  & 82.01  & 82.10  & 82.04  &       & 74.22  & 65.54  & 92.60  & 91.26  & 91.91  \\
          & \textbf{CAT+GPT3.5} & \textbf{51.38} & \textbf{35.55} & \textbf{84.31} & \textbf{85.13} & \textbf{84.70} &       & \textbf{80.82} & \textbf{73.51} & \textbf{95.17} & \textbf{94.64} & \textbf{94.89} \\
          & Role-play+GPT4 & 46.57  & 30.88  & 79.72  & 80.97  & 80.32  &       & 61.68  & 46.38  & 84.95  & 85.11  & 85.01  \\
          & Read+GPT4 & 39.20  & 25.44  & 73.91  & 81.24  & 77.38  &       & 57.37  & 45.60  & 80.62  & 88.26  & 84.24  \\
          & \textbf{CAT+GPT4} & 48.70  & 32.66  & 81.12  & 83.23  & 82.14  &       & 63.68  & 48.24  & 87.04  & 88.26  & 87.63  \\
    \bottomrule
    \end{tabular}}
  \label{tabA5}%
\end{table}%

% Table generated by Excel2LaTeX from sheet 'Sheet3'
\begin{table}[h]
  \centering
  \renewcommand\arraystretch{1.0}
  \caption{Human evaluation results on ``Social Media'' dataset.}
  \resizebox{10cm}{!}{
    \begin{tabular}{ccccc}
    \toprule
    \textbf{Chinese Text} & \textbf{Models} & \textbf{Transfer Accuracy} & \textbf{Content Presevation} & \textbf{Fluency} \\
    \midrule
    \midrule
    \multicolumn{1}{c}{\multirow{18}[4]{*}{\textbf{Social Media}}} & Style Transformer & 59    & 92    & 55  \\
          & RALoCN & 59    & 95    & 58  \\
          & StyleLM & 60    & 94    & 59  \\
          & SDR   & 63    & 94    & 62  \\
          & DRAG  & 64    & 94    & 64  \\
          & StoryTrans & 66    & 94    & 65  \\
\cmidrule{2-5}          & Role-play+ChatGLM & 80    & 88    & 75  \\
          & Read+ChatGLM & 65    & 68    & 70  \\
          & \textbf{CAT+ChatGLM} & 82    & 90    & 80  \\
          & Role-play+Baichuan & 77    & 85    & 65  \\
          & Read+Baichuan & 63    & 66    & 55  \\
          & \textbf{CAT+Baichuan} & 79    & 89    & 75  \\
          & Role-play+GPT3.5 & 81    & 92    & 85  \\
          & Read+GPT3.5 & 83    & 92    & 80  \\
          & \textbf{CAT+GPT3.5} & \textbf{83} & \textbf{95} & 90  \\
          & Role-play+GPT4 & 78    & 78    & 85  \\
          & Read+GPT4 & 78    & 84    & 80  \\
          & \textbf{CAT+GPT4} & 81    & 86    & \textbf{95} \\
    \bottomrule
    \end{tabular}}
  \label{tabA6}%
\end{table}%

% Table generated by Excel2LaTeX from sheet 'Sheet4'
\begin{table}[h]
  \centering
  \renewcommand\arraystretch{1.0}
  \caption{Human evaluation results on ``Legal Document'' dataset.}
  \resizebox{10cm}{!}{
    \begin{tabular}{ccccc}
    \toprule
    \textbf{Chinese Text} & \textbf{Models} & \textbf{Transfer Accuracy} & \textbf{Content Presevation} & \textbf{Fluency} \\
    \midrule
    \midrule
    \multicolumn{1}{c}{\multirow{18}[4]{*}{\textbf{Legal Document}}} & Style Transformer & 59    & 97    & 55  \\
          & RALoCN & 60    & 92    & 58  \\
          & StyleLM & 65    & 95    & 59  \\
          & SDR   & 70    & 94    & 62  \\
          & DRAG  & 75    & 96    & 64  \\
          & StoryTrans & 80    & 96    & 65  \\
\cmidrule{2-5}          & Role-play+ChatGLM & 85    & 88    & 75  \\
          & Read+ChatGLM & 70    & 73    & 70  \\
          & \textbf{CAT+ChatGLM} & 90    & 91    & 80  \\
          & Role-play+Baichuan & 85    & 82    & 70  \\
          & Read+Baichuan & 75    & 74    & 55  \\
          & \textbf{CAT+Baichuan} & 88    & 87    & 80  \\
          & Role-play+GPT3.5 & 90    & 88    & 85  \\
          & Read+GPT3.5 & 85    & 87    & 90  \\
          & \textbf{CAT+GPT3.5} & \textbf{92} & \textbf{94} & 90  \\
          & Role-play+GPT4 & 90    & 86    & 90  \\
          & Read+GPT4 & 88    & 88    & 95  \\
          & \textbf{CAT+GPT4} & 95    & 86    & \textbf{95} \\
    \bottomrule
    \end{tabular}}
  \label{tabA7}%
\end{table}%

% Table generated by Excel2LaTeX from sheet 'Sheet5'
\begin{table}[h]
  \centering
  \renewcommand\arraystretch{1.0}
  \caption{Human evaluation results on ``Scientific Literature'' dataset.}
  \resizebox{10cm}{!}{
    \begin{tabular}{ccccc}
    \toprule
    \textbf{Chinese Text} & \textbf{Models} & \textbf{Transfer Accuracy} & \textbf{Content Presevation} & \textbf{Fluency} \\
    \midrule
    \midrule
    \multicolumn{1}{c}{\multirow{18}[4]{*}{\textbf{Scientific Literature}}} & Style Transformer & 59    & 97    & 55  \\
          & RALoCN & 60    & 92    & 58  \\
          & StyleLM & 65    & 95    & 59  \\
          & SDR   & 70    & 94    & 62  \\
          & DRAG  & 75    & 96    & 64  \\
          & StoryTrans & 80    & 96    & 65  \\
\cmidrule{2-5}          & Role-play+ChatGLM & 85    & 88    & 75  \\
          & Read+ChatGLM & 70    & 73    & 70  \\
          & \textbf{CAT+ChatGLM} & 90    & 91    & 80  \\
          & Role-play+Baichuan & 85    & 82    & 70  \\
          & Read+Baichuan & 75    & 74    & 55  \\
          & \textbf{CAT+Baichuan} & 88    & 87    & 80  \\
          & Role-play+GPT3.5 & 90    & 88    & 85  \\
          & Read+GPT3.5 & 85    & 87    & 90  \\
          & \textbf{CAT+GPT3.5} & \textbf{92} & \textbf{94} & 90  \\
          & Role-play+GPT4 & 90    & 86    & 90  \\
          & Read+GPT4 & 88    & 88    & 95  \\
          & \textbf{CAT+GPT4} & 95    & 86    & \textbf{95} \\
    \bottomrule
    \end{tabular}}
  \label{tabA8}%
\end{table}%

\begin{figure}[!h]
  \centering
  \includegraphics[width=\textwidth]{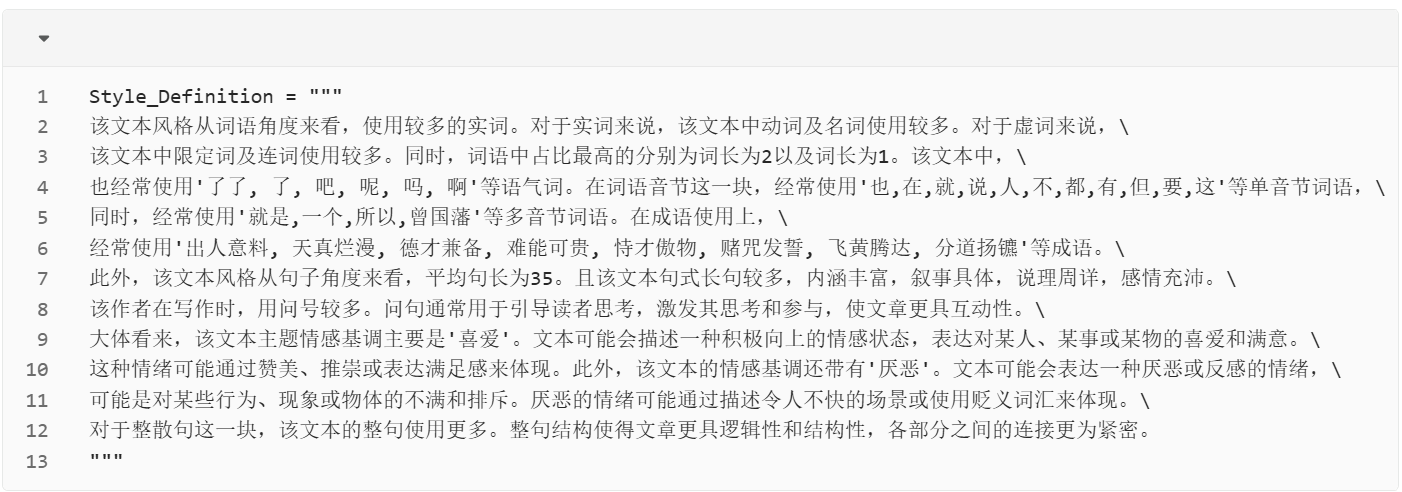}
  \caption{The style definition of ``Fortress Besieged''.}
   \label{figA1}
\end{figure}

\begin{figure}[h]
  \centering
  \includegraphics[width=\textwidth]{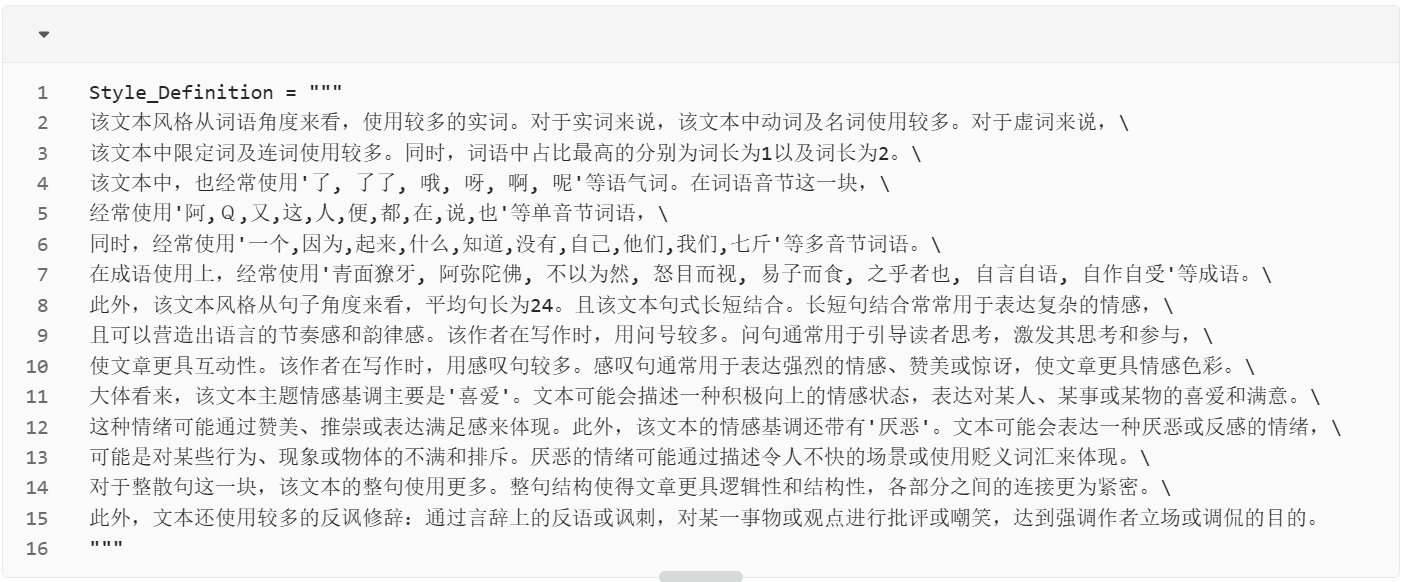}
  \caption{The style definition of ``The Scream''.}
   \label{figA2}
\end{figure}

\begin{figure}[h]
  \centering
  \includegraphics[width=\textwidth]{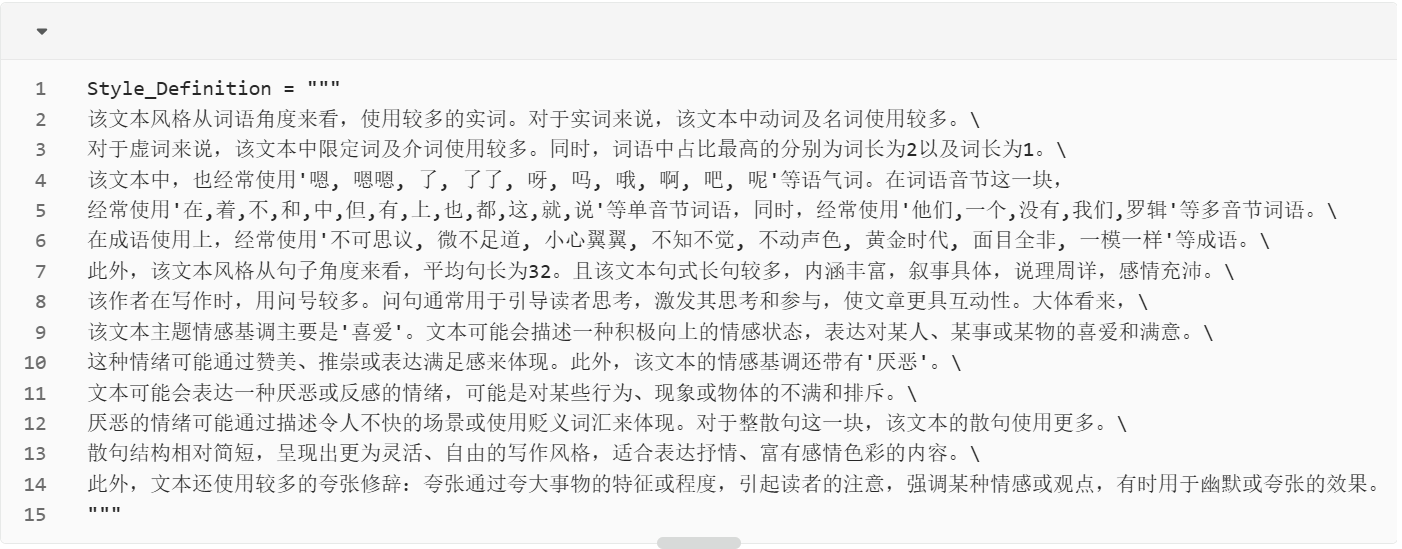}
  \caption{The style definition of ``The Fifteen Year of the Wanli Era''.}
   \label{figA3}
\end{figure}

\begin{figure}[h]
  \centering
  \includegraphics[width=\textwidth]{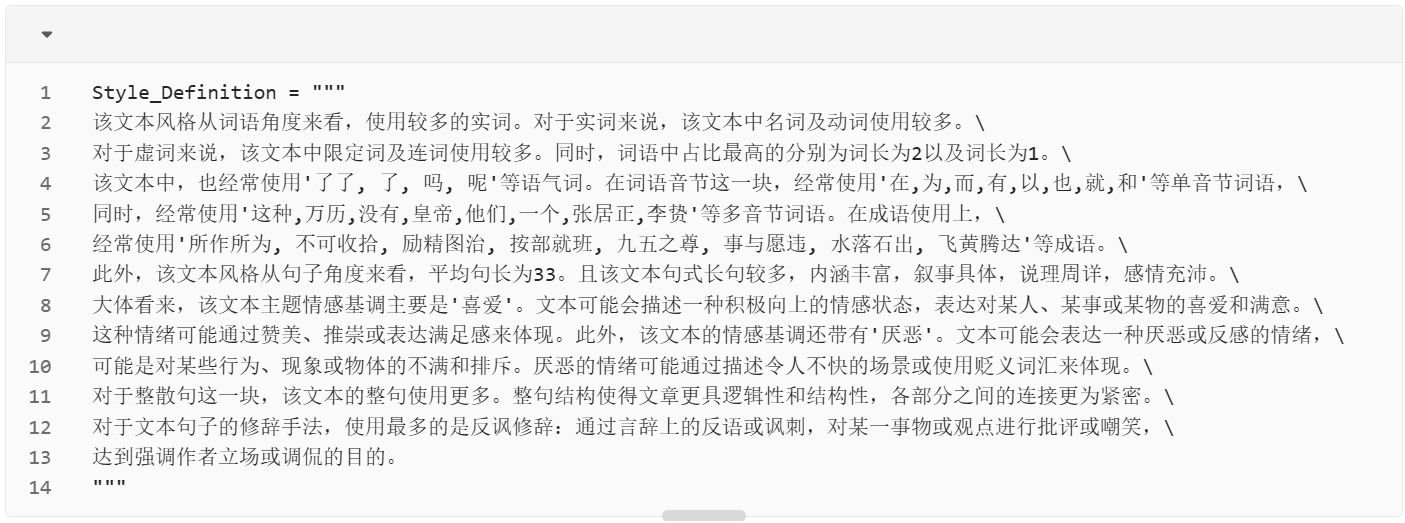}
  \caption{The style definition of ``The Family Instructions of Zeng Guofan''.}
   \label{figA4}
\end{figure} 

\begin{figure}[h]
  \centering
  \includegraphics[width=\textwidth]{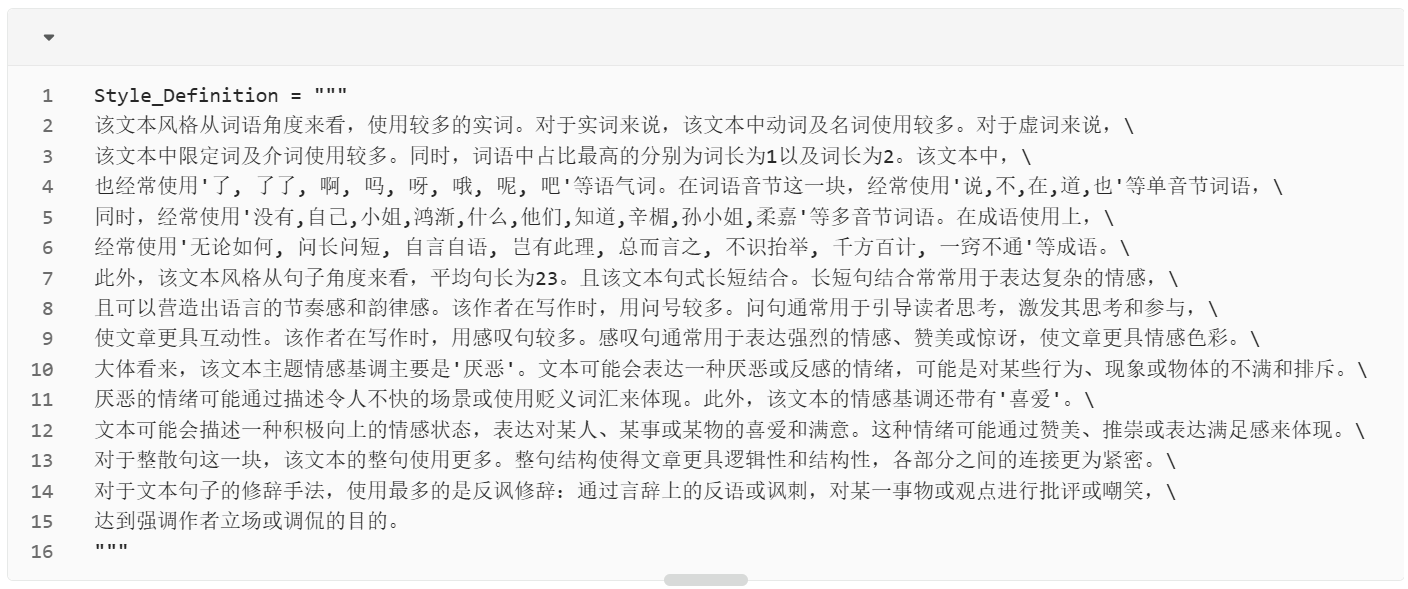}
  \caption{The style definition of ``The Three-Body Problem''.}
   \label{figA5}
\end{figure} 

\begin{figure}[h]
  \centering
  \includegraphics[width=\textwidth]{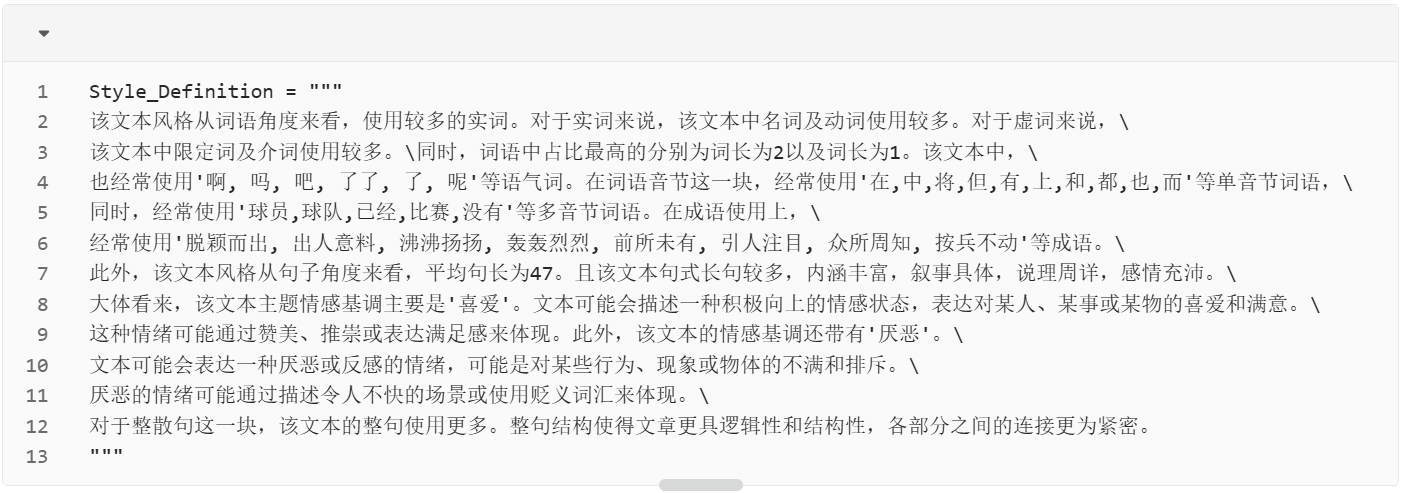}
  \caption{The style definition of ``News'' dataset.}
   \label{figA6}
\end{figure} 

\begin{figure}[h]
  \centering
  \includegraphics[width=\textwidth]{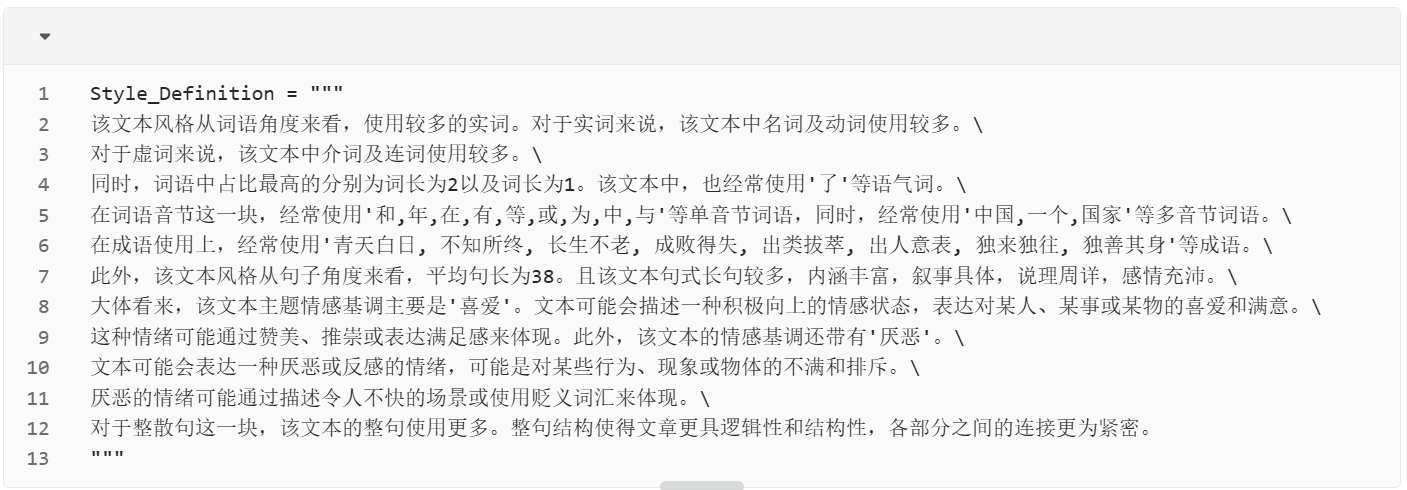}
  \caption{The style definition of ``Wikipedia'' dataset.}
   \label{figA7}
\end{figure} 

\begin{figure}[h]
  \centering
  \includegraphics[width=\textwidth]{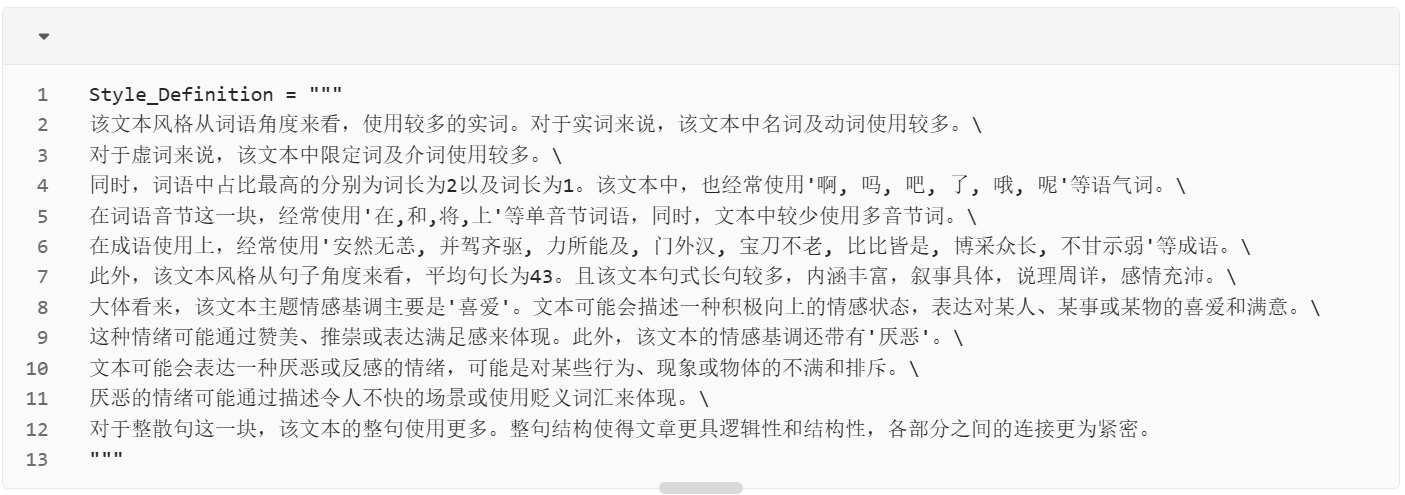}
  \caption{The style definition of ``Social Media'' dataset.}
   \label{figA8}
\end{figure} 

\begin{figure}[h]
  \centering
  \includegraphics[width=\textwidth]{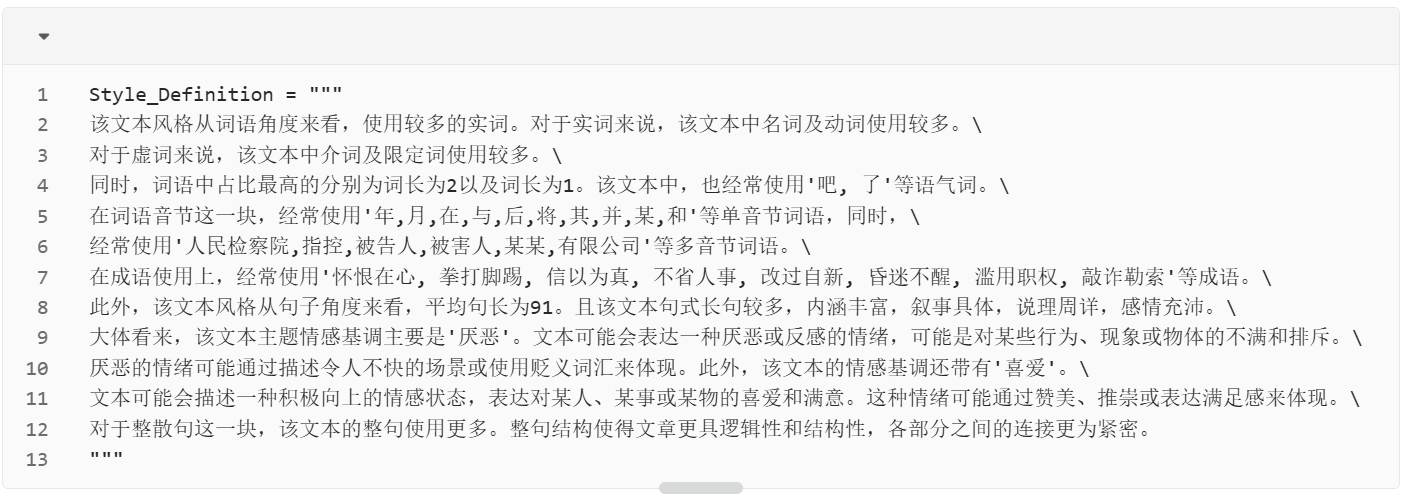}
  \caption{The style definition of ``Legal Document'' dataset.}
   \label{figA9}
\end{figure}

\begin{figure}[h]
  \centering
  \includegraphics[width=\textwidth]{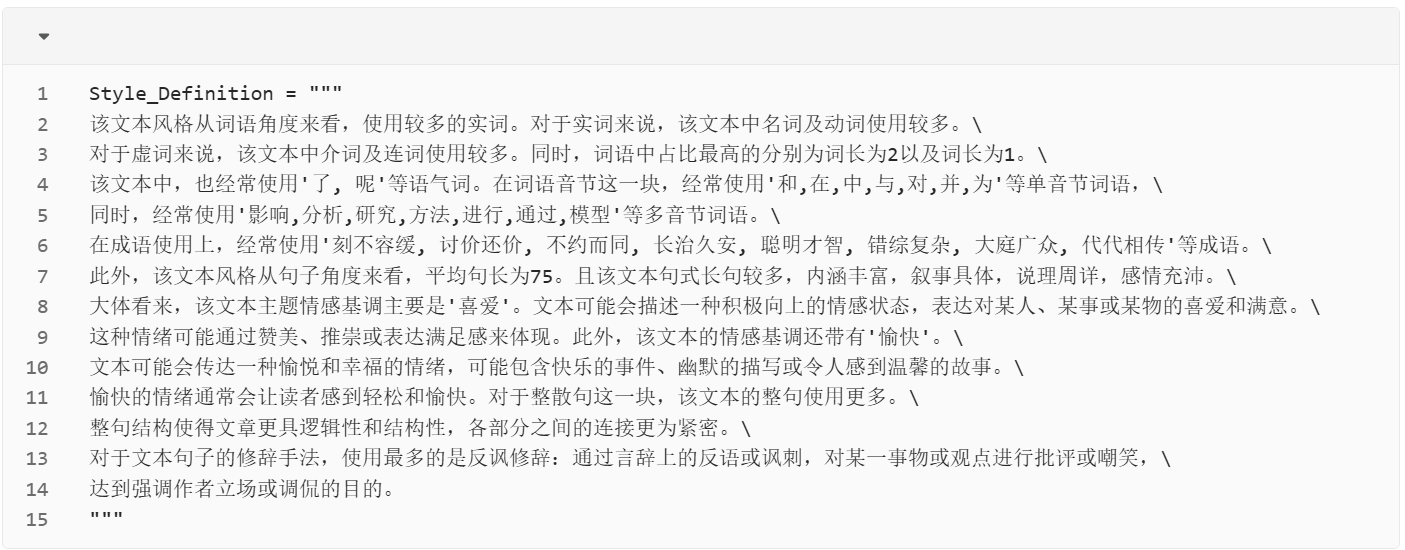}
  \caption{The style definition of ``Scientific Literature'' dataset.}
   \label{figA10}
\end{figure}

\begin{figure}[h]
  \centering
  \includegraphics[width=12cm]{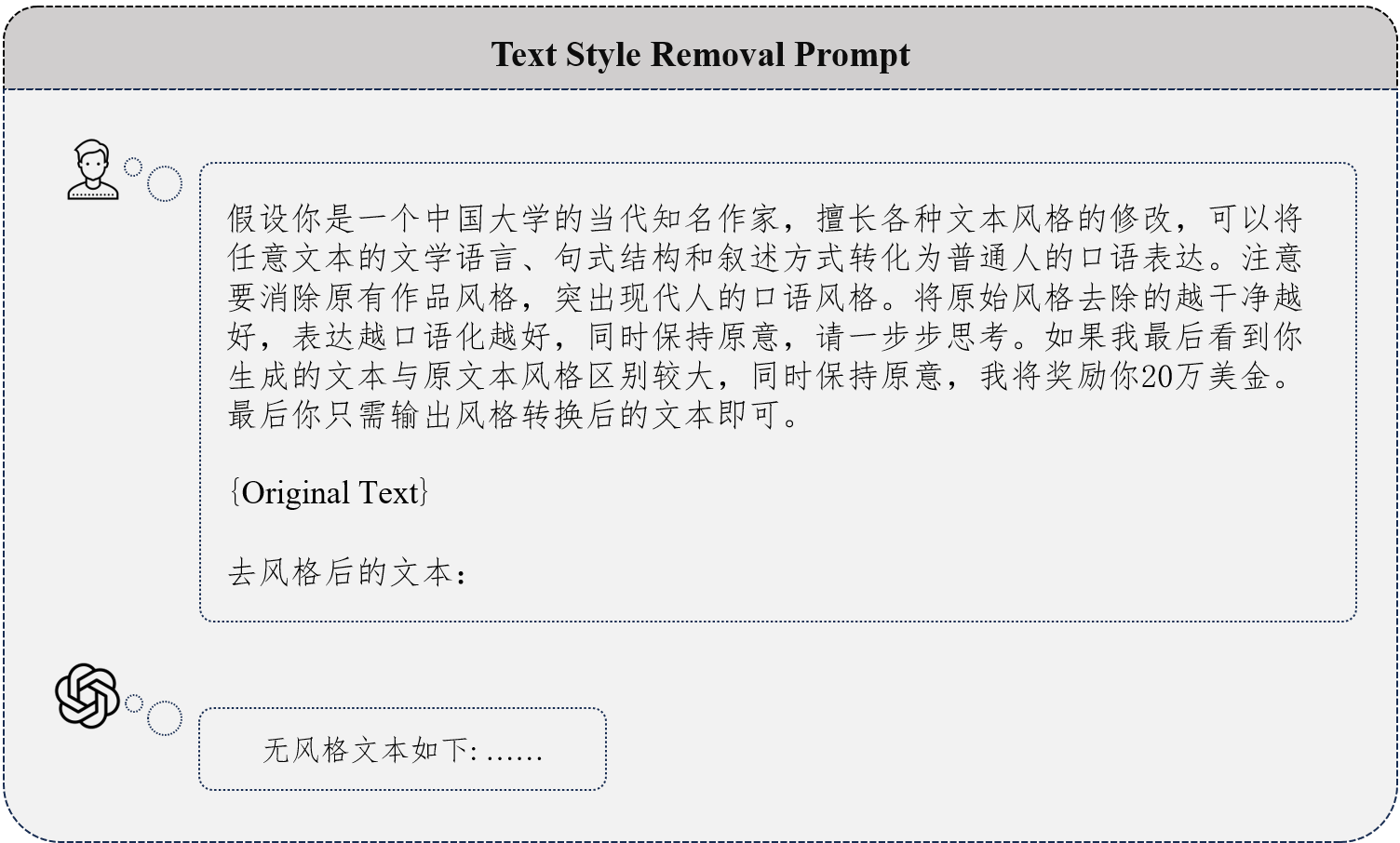}
  \caption{Text style removal prompt.}
   \label{figA11}
\end{figure}

\begin{figure}[h]
  \centering
  \includegraphics[width=12.5cm]{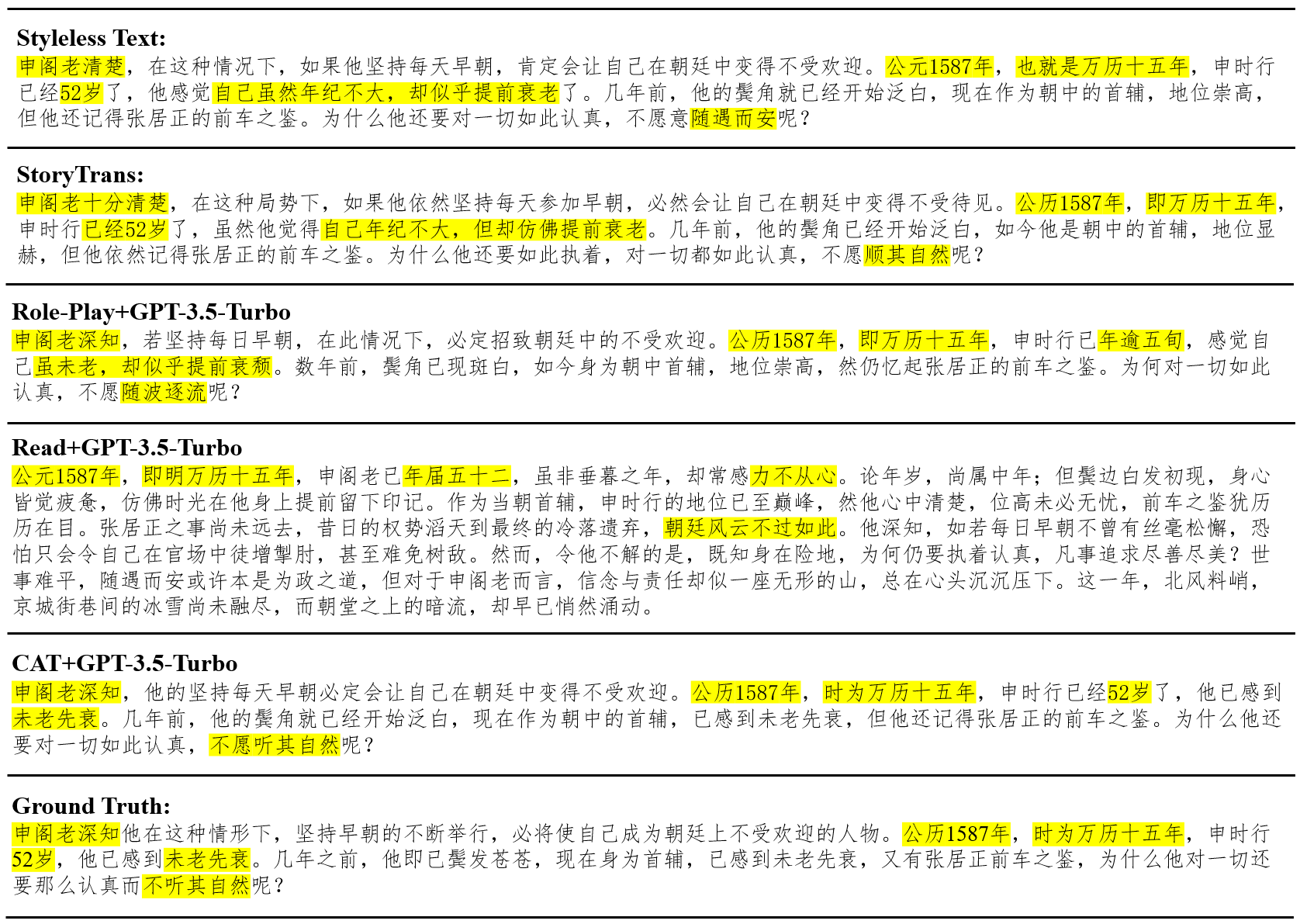}
  \caption{Case study on the ``The Fifteen Year of the Wanli Era'' dataset.}
   \label{figA12}
\end{figure}

\begin{figure}[h]
  \centering
  \includegraphics[width=12.5cm]{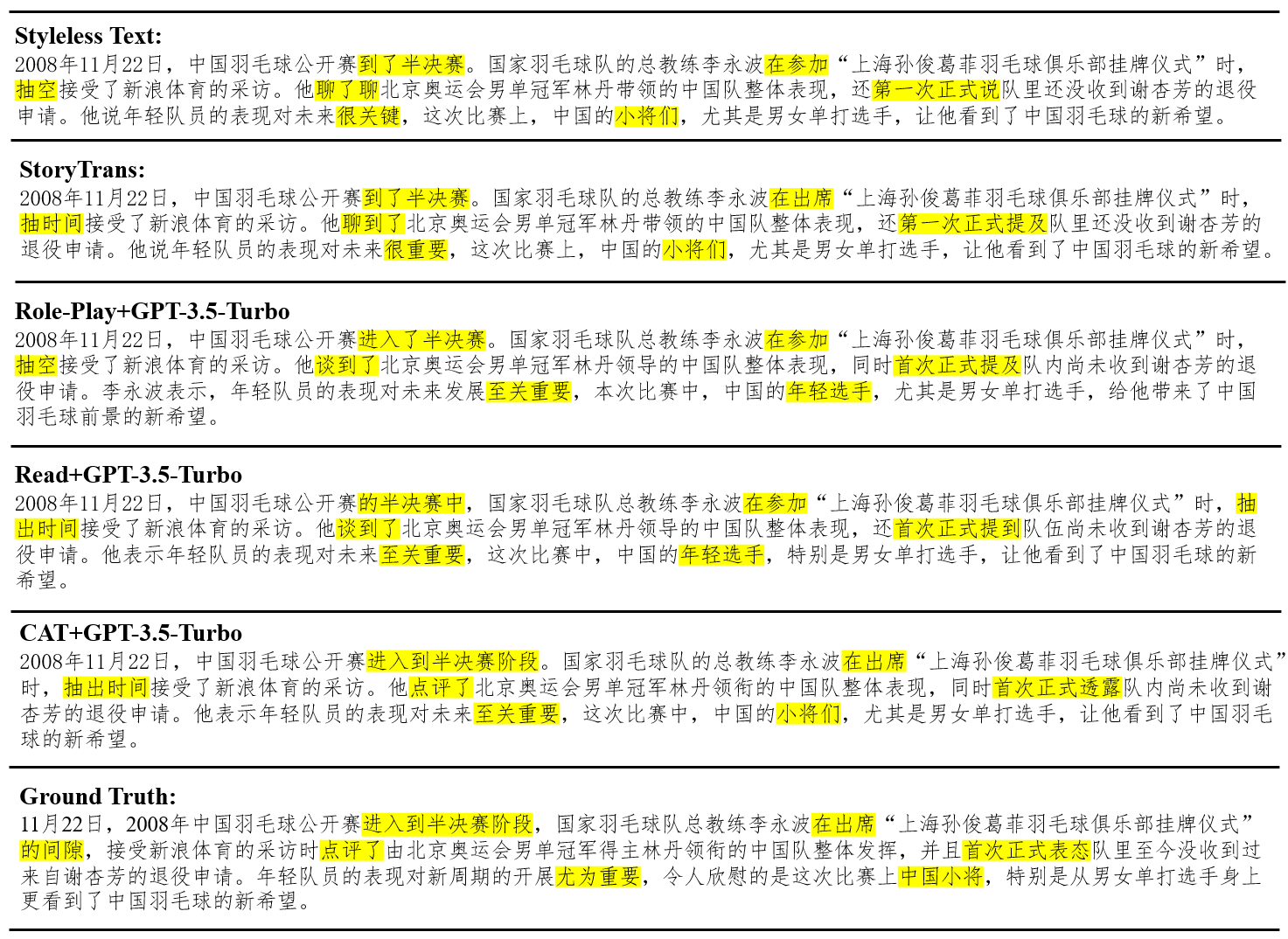}
  \caption{Case study on the ``News'' dataset.}
   \label{figA13}
\end{figure}

\begin{figure}[h]
  \centering
  \includegraphics[width=12.5cm]{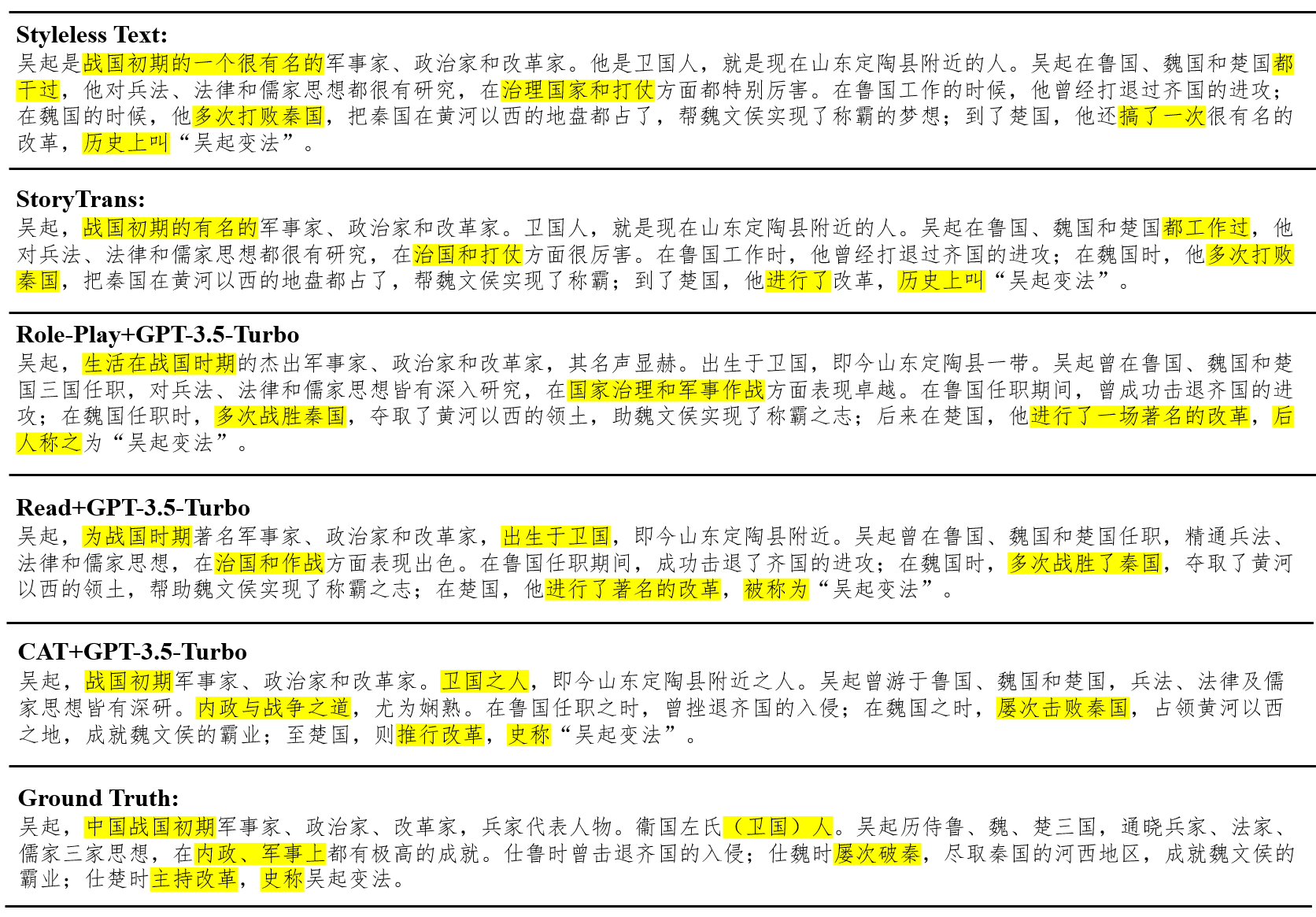}
  \caption{Case study on the ``Wikipedia'' dataset.}
   \label{figA14}
\end{figure}

\begin{figure}[h]
  \centering
  \includegraphics[width=12.5cm]{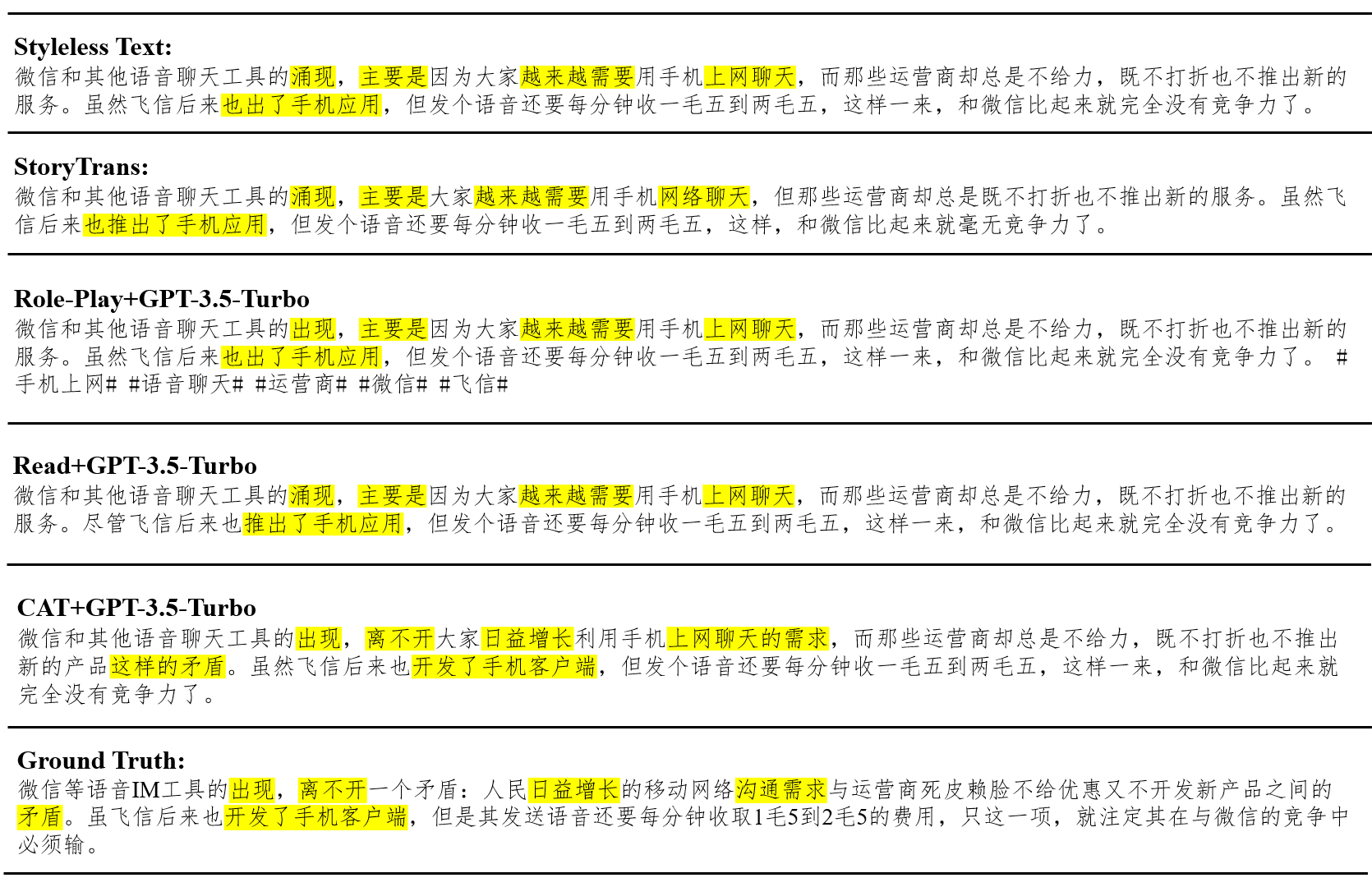}
  \caption{Case study on the ``Social Media'' dataset.}
   \label{figA15}
\end{figure}

\begin{figure}[h]
  \centering
  \includegraphics[width=12.5cm]{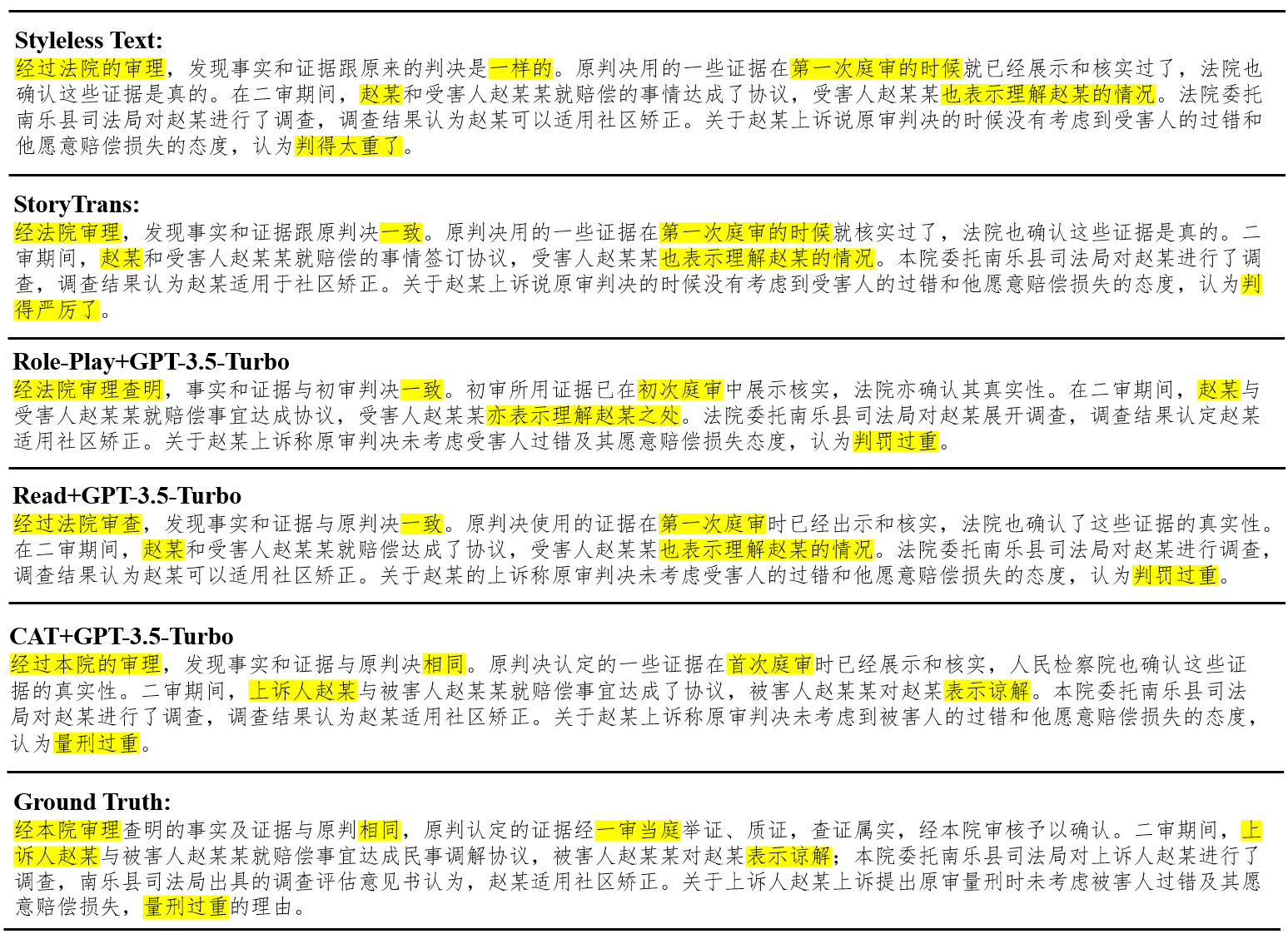}
  \caption{Case study on the ``Legal Document'' dataset.}
   \label{figA16}
\end{figure}

\begin{figure}[h]
  \centering
  \includegraphics[width=12.5cm]{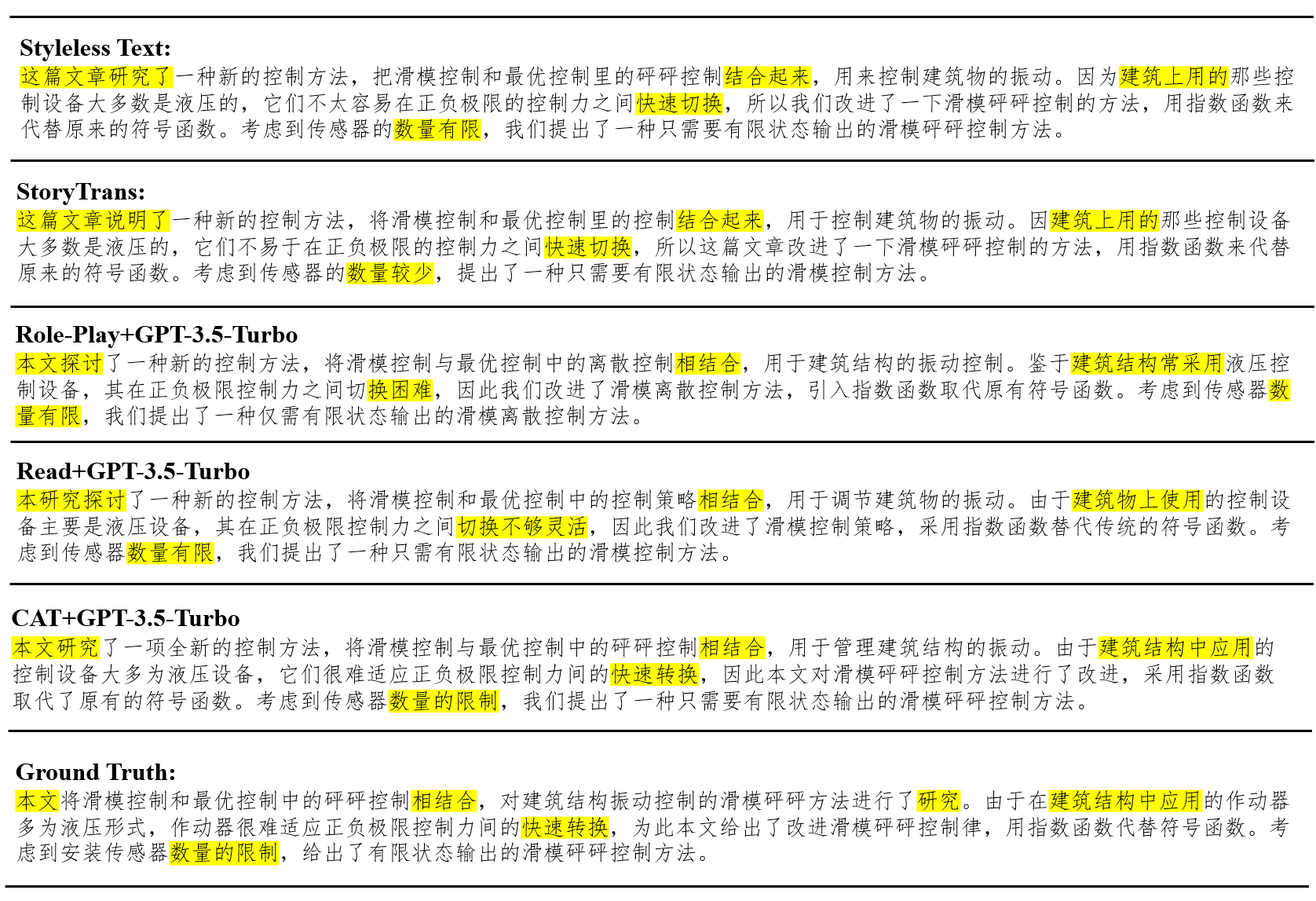}
  \caption{Case study on the ``Scientific Literature'' dataset.}
   \label{figA17}
\end{figure}

\end{document}